\PassOptionsToPackage{dvipsnames,table}{xcolor}
\PassOptionsToPackage{dvipsnames}{colortbl}
\pdfoutput=1

\documentclass[11pt]{article}
\usepackage[table]{xcolor}
\usepackage[dvipsnames]{xcolor,colortbl}

\usepackage[]{acl}

\usepackage{times}
\usepackage{latexsym}
\usepackage{soul}

\usepackage{booktabs}
\usepackage{amsmath}
\usepackage{array}

\usepackage{pifont}
\usepackage{amssymb}
\usepackage{dblfloatfix}
\usepackage{longtable}

\usepackage[T1]{fontenc}

\usepackage[utf8]{inputenc}

\usepackage{microtype}

\usepackage{inconsolata}
\usepackage{multirow}
\usepackage{booktabs}%
\newcommand{\tabitem}{~~\llap{\textbullet}~~}

\usepackage[textsize=scriptsize]{todonotes}

\newenvironment{myquote}[1]%
  {\list{}{\leftmargin=#1\rightmargin=#1}\item[]}%
  {\endlist}

\definecolor{real_teal}{rgb}{27,254,224}

\title{{\sc FactPICO}: Factuality Evaluation for \\ Plain Language Summarization of Medical Evidence}

\author{Sebastian Joseph$^{1}$\footnotemark[1]\ \ \ \
Lily Chen$^{2}$\thanks{Equal Contribution}\ \ \ \ 
Jan Trienes$^{3,4}$\ \ \ \ 
Hannah Louisa Göke$^{3,4}$\\
\textbf{Monika Coers}$^{3,4}$\ \ \ \
\textbf{Wei Xu}$^{5}$\ \ \ \ 
\textbf{Byron C.\ Wallace}$^{6}$\ \ \ \ 
\textbf{Junyi Jessy Li}$^{1}$
\\
$^1$The University of Texas at Austin,
$^2$Massachusetts Institute of Technology\\
$^3$University of Duisburg-Essen,
$^4$Institute for AI in Medicine, University Hospital Essen \\
$^5$Georgia Institute of Technology, $^6$Northeastern University\\
{\small \tt \{sebaj, jessy\}@utexas.edu, l1ly@mit.edu, jan.trienes@uni-due.de} \\
{\small \tt \{hannah.goeke,monika.coers\}@stud.uni-due.de} \\
{\small \tt wei.xu@cc.gatech.edu, b.wallace@northeastern.edu}
}  

\begin{document}
\maketitle
\begin{abstract}
Plain language summarization with LLMs can be useful for improving textual accessibility of technical content.
But how factual are these summaries in a high-stakes domain like medicine? %
This paper presents {\sc FactPICO}, a factuality benchmark for plain language summarization of medical texts describing randomized controlled trials (RCTs), which are the basis of evidence-based medicine and can directly inform patient treatment. {\sc FactPICO} consists of 345 plain language summaries of RCT abstracts generated from three LLMs (i.e., GPT-4, Llama-2, and Alpaca), with fine-grained evaluation and natural language rationales from experts.
We assess the factuality of critical elements of RCTs in those summaries: Populations, Interventions, Comparators, Outcomes (PICO), as well as the reported findings concerning these. %
We also %
evaluate the correctness of the extra information (e.g., %
explanations) added by LLMs.
Using {\sc FactPICO}, we benchmark a range of existing factuality metrics, including the newly devised ones based on LLMs. We find that plain language summarization of medical evidence is still challenging, especially when balancing between simplicity and factuality, and that existing metrics correlate poorly with expert judgments on the instance level.

FactPICO and our code is available at \url{https://github.com/lilywchen/FactPICO}.
\end{abstract}

\begin{figure*}[t]
    \centering
    \includegraphics[width=\linewidth]{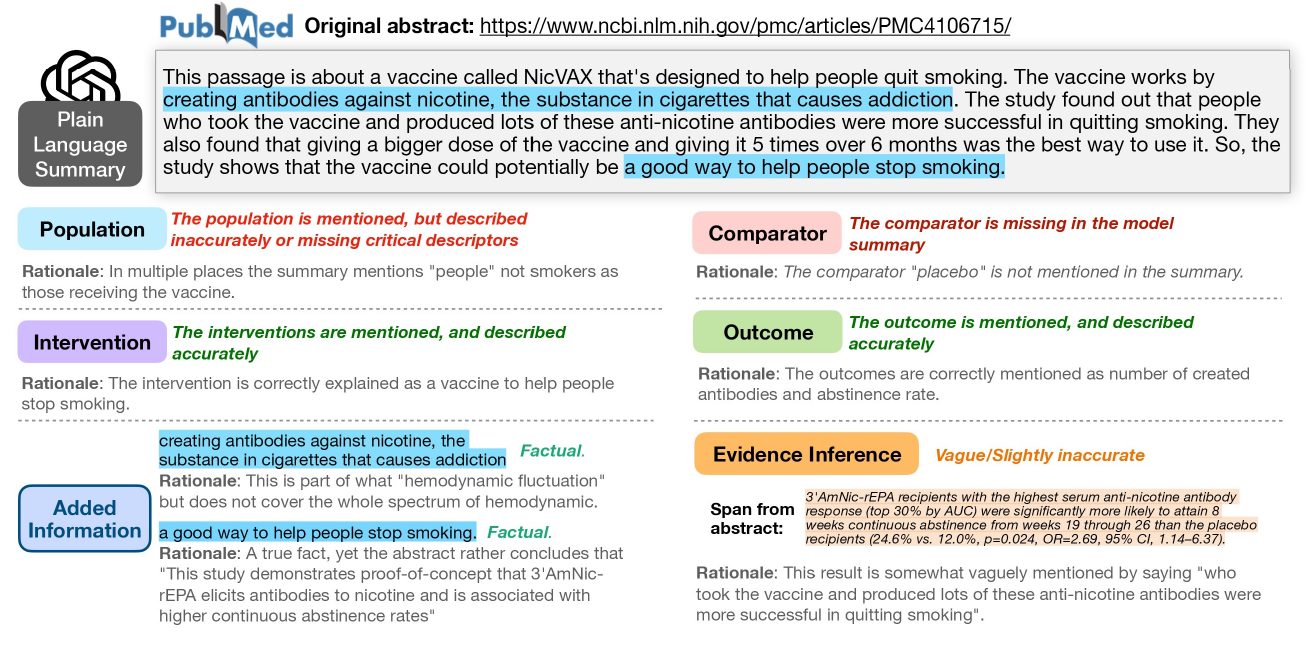} 
    \caption{Expert evaluation of a GPT-4 plain language summary in {\sc FactPICO}. We omitted the original abstract (can be found in Appendix~\ref{sec:abs_full_txt}) in this figure  due to space limit. More examples in Appendix~\ref{app:factpicoexamples}.}
    \label{fig:main_fig2}
\end{figure*}

\section{Introduction}

\label{sec:intro}

New findings in medicine observed in randomized controlled trials (RCTs) are 
published in journal articles which describe their design and outcomes. 
These RCTs ``measure the effectiveness of a new intervention or treatment''~\cite{hariton2018randomised} and 
are the important basis of evidence-based medicine~\cite{sackett1996evidence}. 
However, understanding such articles requires ``specific attention outside of
general literacy capacities''~\cite{american1999health}, rendering them effectively inaccessible to most (lay) people. %
Ideally, healthcare providers would stay current on all medical evidence and share relevant findings with patients, but this is impractical due to the volume
and growth of the evidence base~\cite{bastian2010seventy,marshall2021state}.

LLMs may provide a means for lay readers to access such findings by automatically summarizing and simplifying texts into plain language~\cite{august2023paperplain,shaib-etal-2023-summarizing}. 
Done successfully, this could allow patients to access the most up-to-date literature relevant to their healthcare. 
In turn, this may promote health literacy broadly by disseminating trustworthy information~\cite{Thielmann2023, Cheng2022}. 
But given the inherent risks to personal health, the \emph{factual correctness} of such outputs is paramount in this domain. %

While \citet{shaib-etal-2023-summarizing} showed that GPT-3 
infrequently introduced outright errors when simplifying RCT abstracts, inaccuracies are occasionally introduced; this ought to be addressed before wide adoption of such technology. 
Unfortunately, there is no standard evaluation benchmark for factuality on this important \emph{medical evidence text simplification}
task. 
Consequently, it is %
unknown whether and to what degree existing automatic factuality evaluation metrics align with human judgments.

We posit that focusing on critical elements in the RCT structure is %
key for factual medical evidence communication. This work presents {\sc FactPICO}, an expert-constructed factuality benchmark for the plain language summarization of technical abstracts describing RCTs.\footnote{We focus on abstracts, because they are always publicly accessible, and typically include the key results that would be of interest to individuals.}
{\sc FactPICO} is a fine-grained benchmark %
focused on key characteristics of trials: \textbf{P}opulations 
(e.g.\ \emph{COVID patients; diabetics}),
\textbf{I}nterventions 
(e.g.\ \emph{remdesivir}),
\textbf{C}omparators 
(e.g.\ \emph{placebo}),
\textbf{O}utcomes
(e.g.\ \emph{30 day mortality, or pain}),
as well as  Evidence Inference (i.e., whether the intervention yielded a significant difference in the treated group with respect to the outcome; \citealt{lehman2019inferring}). PICO is a %
standard framework to structure clinical %
questions.

Figure~\ref{fig:main_fig2} shows an example of {\sc FactPICO} annotation. %
In contrast to standard summaries, plain language summaries have the additional goal of simplifying content for lay readers. 
This may involve elaboration and explanation of difficult concepts to foster understanding~\cite{srikanth-li-2021-elaborative}. 
Thus, {\sc FactPICO} includes a correctness assessment of \textbf{added content}. 
{\sc FactPICO} is also distinct in that it includes \textbf{expert-written rationales} that contextualize the evaluation of these fine-grained characteristics, providing a useful first step in assessing explainable factuality evaluation methods.

{\sc FactPICO}
includes outputs from a mix of proprietary and open-source LLMs (GPT-4, Llama-2-Chat, and Alpaca). 
Our findings are somewhat less optimistic than prior work in %
medical~\cite{shaib-etal-2023-summarizing} and news summarization~\cite{goyal2022news}:  
Factual errors (occasionally important ones) with respect to key RCT elements are 
introduced by LLMs. 
One concerning phenomenon is the extent to which models overgeneralize, resulting in problematic information loss~\cite{Trienes:2024:arXiv}.

Using {\sc FactPICO}, we evaluated a suite of existing automatic metrics shown to perform well for factuality in summarization~\cite{scialom2021questeval,goyal2021annotating,fabbri-etal-2022-qafacteval,zha2023alignscore,tang-etal-2023-understanding}, as well as newly devised LLM-based evaluations.
We find that existing metrics correlate with expert ratings at the system level, but not at the instance level. 
The best performing metric is an LLM-based %
approach in which we first identify key RCT elements; this shows that providing models with explicit domain knowledge may help.
Analysis of LLM-generated rationales shows that LLMs often provide flawed reasoning when justifying their self-evaluations.

\section{{\sc FactPICO} Benchmark}

The {\sc FactPICO} benchmark consists of expert factuality assessments of 345 LLM-generated plain language summaries of 115 RCT abstracts.

\subsection{LLM-based Plain Language Summaries}
\label{sec:dataset}

The medical abstracts used in {\sc FactPICO} are sourced from the \textit{Evidence Inference 2.0} dataset \citep{deyoung2020evidence}, 
which contains abstracts and full articles that describe RCTs from PubMed.
We use a subset of the abstracts which include annotated spans that state clinical results.
We also exclude abstracts that have corresponding human-written plain language summaries in the PubMed database, which may have appeared in the pre-training data of the LLMs we are experimenting with.\footnote{We check this by querying the Entrez database system and filtering out abstracts which have other abstracts associated with their PubMed ID.}
{\sc FactPICO} includes 115 abstracts randomly sampled from this filtered subset.

For each abstract, we generate plain language summaries using three different LLMs under zero-shot prompting.
This zero-shot setting better emulates how lay users would likely interact with models, compared to few-shot. 
We use GPT-4 \cite{openai2023gpt4}, Llama-2-Chat \cite{touvron2023llama}, and Alpaca
\cite{alpaca}, which resulted in 345 ($115\times 3$) total outputs.
In a preliminary investigation, we found that these models seemed qualitatively best at generating plain language summaries. 
Details about this pilot study, prompt exploration, and the final prompts are in Appendix \ref{sec:prompts}.

\subsection{Human Evaluation Framework}
\label{fpico_framework}
We evaluate generated summaries using a set of questions addressing factuality as related to the key PICO aspects of RCTs, as well as information added by LLMs during simplification. 
We ask evaluators to \textbf{score} model outputs and provide \textbf{natural language rationales} to justify their ratings. 
Annotation interface details are in Appendix~\ref{sec:interface}.

\paragraph{PICO Elements.}
Population, Intervention, Comparator, Outcome (PICO) elements are the key components of an RCT \cite{Richardson1995-sz}. The trial \textbf{population} concerns characteristics of subjects in the trial, including what condition they have, 
the number of participants, %
and their demographics.  %
The \textbf{intervention} is the active treatment being %
assessed for comparative efficacy; 
the \textbf{comparator} is the control 
with respect to which %
this is being compared. 
Finally, \textbf{outcomes} are those things that are measured to determine results. 

Accurate representation of these essential descriptors in the plain language summaries of an RCT is imperative. 
More explanations and examples are provided in Appendix Table \ref{table:critcial}. 
We ask evaluators to provide a rating between 1 to 4 codifying the factuality with respect to each PICO element, respectively:
\textbf{4:} \textit{Mentioned and described accurately}; 
\textbf{3:} \textit{Mentioned but described somewhat inaccurately or vaguely};
\textbf{2:} \textit{Mentioned but described with severe inaccuracies and/or is missing critical descriptors};
\textbf{1:} \textit{Missing}.
We describe the rating system in greater detail in Appendix \ref{sec:eval_guide}.

\paragraph{Rationales.} Evaluators are also asked to provide natural language \emph{rationales} justifying their chosen rating. Such rationales may reveal technical justifications for annotations which would be difficult for a layperson to assess (see Figure~\ref{fig:span_ex} for examples). More usefully, rationales can express a degree of uncertainty in evaluations. 
For example, consider the following rationale for an intervention being evaluated as accurately mentioned:
\begin{myquote}{0.1in}
    \small
    The interventions are clearly described as one group receiving morphine and one triamcinolone, yet their combination with bupivacaine is missing, but doesn't seem very important.
\end{myquote}
This rationale indicates that the choice to evaluate the intervention as accurate was not a clear-cut decision. By including rationales in {\sc FactPICO}, this complex decision process is documented. We compare LLM rationales and expert rationales in Section~\ref{sec:prelim_rat}, and envision future work to dive deeply into explainable factuality measures.

\paragraph{Evidence Inference.}
We additionally evaluate the \emph{evidence inference} aspect of LLM-generated plain language summaries, i.e., whether the results and findings concerning PICO are reported factually, as PICO covers what outcomes were \emph{measured}, it does not account for the corresponding \emph{results} (e.g., if an intervention was found to outperform a comparator). 
A challenge here is that most trials will report multiple results, any of which may be conveyed (un)factually in the summary. 
Thus, we collect factuality assessments at the level of individual results, using findings annotated in the source abstracts from the \textit{Evidence Inference 2.0} dataset as reference.
Evaluators are asked to determine how well each particular inference is reflected in the plain language summary (\textbf{4:} \textit{accurate}, \textbf{3:} \textit{vague/slightly inaccurate}, \textbf{2:} \textit{inaccurate}, \textbf{1:} \textit{not mentioned}), with a free-text rationale. %

\paragraph{Added Information.} %
Unlike traditional summarization where content addition is seen as extrinsic hallucination~\cite{maynez2020faithfulness}, plain language summarization often requires the model to explain and elaborate complex concepts~\cite{srikanth-li-2021-elaborative}.
It is important to verify the correctness of content additions. 
We ask annotators to highlight addition spans, determine whether they are factual, and justify each rating with a free-text rationale.

\subsection{Annotation}

{\sc FactPICO} was annotated by two senior students in their fifth year of medical school who are highly proficient in English. They are experienced with data annotation for text simplification and summarization tasks.
To %
ensure high annotation quality, we conducted a training phase which involved annotating a set of summaries as pilot (excluded from {\sc FactPICO}).
Next, we collected two sets of annotations on 75 summaries (25 from each model). For the first 60 of these summaries, we asked annotators to discuss their responses with respect to the PICO elements. These discussions were %
intended to resolve %
any conceptual differences related to PICO annotation (Appendix~\ref{sec:discussion}).
The remaining 15 summaries were evaluated individually. 
Compensation was 12€ per hour, which is standard for fixed-hour contracts with the annotators' institutions.

\begin{table}[t]
\small 
    \centering
    \begin{tabular}{lc|lc}
    \toprule
    \textbf{Type} & \textbf{$\kappa$} & \textbf{Type} & \textbf{$\kappa$} \\  \midrule
        Population & 0.56 &  Evidence Inference &  0.47  \\
        Intervention & 0.80 & Added Info & 0.65 \\
        Comparator & 0.73 & \ \ \ - correctness & 0.86 \\
        Outcome & 0.60 \\
        \bottomrule
    \end{tabular}
    \caption{Inter-evaluator agreement.
    For PICO evaluations (left) we show agreement  on the subset of 15 \emph{undiscussed} texts (see Appendix~\ref{sec:discussion}). For other {\em undiscussed} questions (right), we show agreement on all 75 doubly annotated texts.
    }
    \label{table:agg}
\end{table} 
\newcommand{\Example}[5]{
    \textbf{#1} \newline
    \textbf{Source:} #2 \newline
    \textbf{Simplification [#3]:} #4 \newline
    \textbf{Rationale:} #5
}
\begin{table*}[t]
    \centering
    \scriptsize
\begin{tabular}{
p{0.97\textwidth}} \toprule
        \Example{Supplementary Elaboration \textcolor{teal}{(factual)}}{We conducted a randomized, placebo-controlled clinical trial to prove efficacy of botulinum toxin for neck pain in chronic whiplash syndrome.}{GPT-4}{Scientists did a study to see if a certain type of medicine, called botulinum toxin, could help people who had long-lasting neck pain from whiplash (\underline{an injury that happens if your head jerks hard}).}
         {
            This is an explanation that is not mentioned in the original.
         }\vspace{0.3em}
         \\
         \rowcolor{gray!10}
         \Example{Substituting Elaboration \textcolor{teal}{(factual)}}{The aim of the study was to compare the effect of co-administered dexamethasone and diclofenac potassium (diclofenac K) with diclofenac K alone on the postoperative pain, swelling and trismus after surgical removal of third molars.}{Llama-2}{
            The study compared how well two different medicines, dexamethasone and diclofenac K, worked together to help people feel better after having surgery to remove their \underline{wisdom teeth}. 
        }{
            This is the correct explanation for 3rd molar.
        }
        \vspace{0.3em}
         \\
         \Example{Nonsensical Hallucination \textcolor{red}{(non-factual)}}{Intra-articular morphine was as effective as intra-articular triamcinolone for analgesia in patients with osteoarthritis knee. Intra-articular morphine is possibly a better option than intra-articular steroid as it has lesser side effects.}{GPT-4}{The study concluded that both morphine and triamcinolone are useful, but morphine might be better as it has fewer side effects. \underline{So, if granny or grandpa is having pain because of osteoarthritis, an injection with morphine could help them better.}}{
            This is definitely not factual or good advice to give, even though the information within is technically true.
         }\vspace{0.3em}
         \\
         \rowcolor{gray!10}
         \Example{Jumping to Conclusions \textcolor{red}{(non-factual)}}{This study aimed to determine if raloxifene prevents the negative effects of diabetes on skeletal fragility in diabetes-prone rats.}{Llama-2}{\underline{Raloxifene is a medicine that can help prevent diabetes from causing weak bones.}}{
             The abstract says that this study aims to test IF Raloxifene can help prevent diabetes from causing "weak bones"/"if raloxifene prevents the negative effects of diabetes on skeletal fragility in diabetes-prone rats"
         }\\
    \bottomrule
\end{tabular}
\vspace{-0.5em}
\caption{Examples of added information (\underline{underlined}) found in plain language summaries within {\sc FactPICO}. 
}
\label{fig:span_ex}
\end{table*}

\paragraph{Inter-Evaluator Agreement.} 
Table~\ref{table:agg} reports agreement on the 15 held-out (undiscussed) subset for PICO elements, as well as agreement on the full set of 75 texts for the Evidence Inference and Added Information questions (all annotated without discussion). We report Randolph's kappa~\cite{randolph2005free}, a free-marginal version of Fleiss' kappa.\footnote{Free-marginal refers to distributions where raters have no prior knowledge as to the quantity of instances that would be assigned each rating, such as is the case with {\sc FactPICO}.}
With respect to added information, the agreement reported in Table~\ref{table:agg} is at the sentence level (for each sentence, whether it is considered as added information). The correctness of added information is reported on the 48 spans both annotators identified as added information. 
Overall, evaluators showed moderate to high  agreement~\cite{artstein2008inter}. 
This range of kappa values are expected given %
the degree of subjectivity %
inherent to rating the severity of factual errors.

\begin{table*}[t]
\centering
\small
\begin{tabular}{l|rrrrrr|rrrrr}
\toprule
{} &  \textbf{Pop.} &  \textbf{Inter.} &  \textbf{Comp.} &  \textbf{Out.}  &  \textbf{E.Inf.} &  \textbf{Average} & \textbf{\#tokens} & \textbf{$\Delta$FK$\uparrow$} & \textbf{Rg-L$\downarrow$}  & \textbf{\#N$\downarrow$} & \textbf{\%N$\downarrow$}\\
\midrule
ALPACA     & \textbf{3.30} &   \textbf{3.70} &  \textbf{3.42} & \textbf{3.77}  &    \textbf{3.46} & \textbf{3.53} & 170.95 & -0.61 & 0.479 & \textbf{8}  & \textbf{7.0}\\
GPT-4      & 3.12 &   3.52 &  3.20 & 3.56 &    3.25 & 3.33 & 162.91 & 2.87 & 0.146 & 53 & 31.3\\
LLAMA-2    & 2.71 &   3.40 &  2.70 & 3.41  &    2.80 & 3.00 & 116.27 &  \textbf{2.92} & \textbf{0.136} & 57 & 38.3\\
\bottomrule
\end{tabular}
\vspace{-0.5em}
\caption{Human evaluation on a 1-4 scale $\uparrow$ (Section \ref{fpico_framework}) on the factuality of PICO elements and Evidence Inferences. The average length of the original technical abstracts is 343.5 tokens. \textbf{$\Delta$FK} and \textbf{Rg-L}: Flesch Kincaid Grade Level difference and ROUGE-L between abstract and summary. \textbf{\#N}: number of non-factual additions. \textbf{\%N}: percentage of summaries in {\sc FactPICO} with at least one non-factual addition. }
\label{table:llm_analysis}
\end{table*}
\section{Evaluation and Analysis of Plain Language Summaries}
Results of the factuality evaluation of the plain language summaries generated by GPT-4, Llama-2, and Alpaca in {\sc FactPICO} are presented in Table \ref{table:llm_analysis}. 
We report average ratings for each PICO category and Evidence Inference, as well as the number and percentage of non-factual added information spans identified by either of the evaluators. 
Since readability is a key goal of {\em plain language} summarization, we also report the change in Flesch-Kincaid Grade Level \cite{fkgl} between the abstract and the summary ($\Delta$FK),
as well as the ROUGE-L score~\cite{lin-2004-rouge}, which quantifies $n$-gram overlap between the generated plain language summary and the original abstract (i.e., high overlap may indicate low readability).

\paragraph{Factuality vs. Simplicity.}
We observe a clear trade-off between the simplicity of the generated text and its factuality. 
Plain language summaries generated by Alpaca are rated as more factual across all dimensions, including most of the automatic metrics (Table~\ref{table:grade}) discussed in Section~\ref{sec:autoevalresults}. However, Alpaca is also the most extractive (i.e., heavily relying on deletion), with a near 0.5 ROUGE-L score and an %
advanced reading level.
By contrast, GPT-4 and Llama-2 both simplify by rephrasing, with similar ROUGE-L and reading levels. Yet the plain language summaries generated by GPT-4 and Llama-2 are less factual, with a significant increase in the number of hallucinations (added non-factual information). Comparatively,
Llama-2 produced the least factual summaries.

\paragraph{Can LLMs reliably convey \emph{critical} RCT elements?}
The fine-grained framework in {\sc FactPICO} exposes issues with LLM-generated plain language summaries for medical texts. As shown in Table~\ref{table:llm_analysis}, LLMs can explain the Interventions and Outcomes more accurately, while failing to do so for other information, such as the Populations, Comparators, and Evidence Inference.

Overgeneralization and omissions of such critical elements may distort the findings and conclusions of medical research studies in nuanced but important ways. Consider this example where GPT-4 omitted a critical element of RCT in its summary.

\begin{myquote}{0.1in}
\small
Scientists did a study to see if people who know they have a gene that puts them at risk for heart problems would eat healthier. They looked at people's intake of a vitamin called folate, which is good for heart health. The participants in the study were told if they have the risk gene or not. But in the end, there was no difference in how much folate the people with the risk gene ate compared to those without the risk gene. So, knowing if they have the risk gene didn't make people eat healthier.
\end{myquote}
In the original RCT abstract, participants received ``either general healthy eating advice or varying levels of personalised nutrition advice''. 
One of the levels of such personalized advice involved testing for a gene ``in relation to cardiovascular health and the importance of a sufficient intake of folate''.
But, the generated summary above fails to mention the comparator, which is the general healthy eating advice; instead, it focuses %
on the overgeneralized intervention, i.e., personalized advice involving genetic variants (without mentioning %
the personalized nutrition advice). 
These omissions not only %
render the simplification incomplete, but they also make understanding the actual result impossible.

\vspace{-0.5em}
\paragraph{Do LLMs generate accurate elaborations?}
Table~\ref{table:llm_analysis} also shows a concerning amount of non-factual additions within the generated simplifications. Examples of these span-level annotations can be found in Table~\ref{fig:span_ex}. 
Most of these non-factual additions may misrepresent the original medical text and consequently mislead lay readers. %
Even GPT-4
produced many plain language summaries (31.3\% of all additions added) with such errors, %
raising questions as to the trustworthiness of LLMs 
for tasks in high-stakes domains such as medicine, when used by zero-shot prompting (a most common use-case for lay users).

\begin{table*}[t]
\small
\centering
\begin{tabular}{lrrrrrrrrrr}
\toprule
 & & \textbf{QAFact} & \textbf{Quest} & \textbf{AlignS} & \textbf{DAE} & \textbf{GPT-4} & \textbf{Llama-2} & \textbf{Alpaca} & \textbf{Mistral} & 
 \textbf{Extract} \\
\specialrule{.4pt}{2pt}{0pt}
\multirow{6}{*}{$\tau_{b}$} & Pop.  & {\cellcolor[HTML]{00692A}} \color[HTML]{F1F1F1} 0.237 & {\cellcolor[HTML]{087432}} \color[HTML]{F1F1F1} 0.226 & {\cellcolor[HTML]{218944}} \color[HTML]{F1F1F1} 0.201 & {\cellcolor[HTML]{84CC83}} \color[HTML]{000000} 0.116 & {\cellcolor[HTML]{00491D}} \color[HTML]{F1F1F1} 0.265 & {\cellcolor[HTML]{D6EFD0}} \color[HTML]{000000} 0.042 & {\cellcolor[HTML]{ABDDA5}} \color[HTML]{000000} 0.084 & {\cellcolor[HTML]{F7FCF5}} \color[HTML]{000000} -0.011 & {\cellcolor[HTML]{00441B}} \color[HTML]{F1F1F1} 0.270 \\
& Inter. &    {\cellcolor[HTML]{00451C}} \color[HTML]{F1F1F1} 0.232 & {\cellcolor[HTML]{0B7734}} \color[HTML]{F1F1F1} 0.202 & {\cellcolor[HTML]{A2D99C}} \color[HTML]{000000} 0.117 & {\cellcolor[HTML]{91D28E}} \color[HTML]{000000} 0.126 & {\cellcolor[HTML]{005622}} \color[HTML]{F1F1F1} 0.223 & {\cellcolor[HTML]{EFF9EB}} \color[HTML]{000000} 0.059 & {\cellcolor[HTML]{F7FCF5}} \color[HTML]{000000} 0.048 & {\cellcolor[HTML]{C8E9C1}} \color[HTML]{000000} 0.093 & {\cellcolor[HTML]{00441B}} \color[HTML]{F1F1F1} 0.234 \\
& Comp. & {\cellcolor[HTML]{8BCF89}} \color[HTML]{000000} 0.176 & {\cellcolor[HTML]{8BCF89}} \color[HTML]{000000} 0.177 & {\cellcolor[HTML]{AEDEA7}} \color[HTML]{000000} 0.140 & {\cellcolor[HTML]{BBE4B4}} \color[HTML]{000000} 0.123 & {\cellcolor[HTML]{006D2C}} \color[HTML]{F1F1F1} 0.341 & {\cellcolor[HTML]{EFF9EC}} \color[HTML]{000000} 0.036 & {\cellcolor[HTML]{DDF2D8}} \color[HTML]{000000} 0.073 & {\cellcolor[HTML]{F7FCF5}} \color[HTML]{000000} 0.015 & {\cellcolor[HTML]{00441B}} \color[HTML]{F1F1F1} 0.387 \\
& Out. &  {\cellcolor[HTML]{137D39}} \color[HTML]{F1F1F1} 0.228 & {\cellcolor[HTML]{238B45}} \color[HTML]{F1F1F1} 0.214 & {\cellcolor[HTML]{218944}} \color[HTML]{F1F1F1} 0.216 & {\cellcolor[HTML]{95D391}} \color[HTML]{000000} 0.130 & {\cellcolor[HTML]{00441B}} \color[HTML]{F1F1F1} 0.276 & {\cellcolor[HTML]{D3EECD}} \color[HTML]{000000} 0.078 & {\cellcolor[HTML]{F7FCF5}} \color[HTML]{000000} 0.028 & {\cellcolor[HTML]{CFECC9}} \color[HTML]{000000} 0.082 & {\cellcolor[HTML]{005120}} \color[HTML]{F1F1F1} 0.266 \\
& Evd.Inf & {\cellcolor[HTML]{4EB264}} \color[HTML]{F1F1F1} 0.248 & {\cellcolor[HTML]{6BC072}} \color[HTML]{000000} 0.221 & {\cellcolor[HTML]{55B567}} \color[HTML]{F1F1F1} 0.242 & {\cellcolor[HTML]{E2F4DD}} \color[HTML]{000000} 0.075 & {\cellcolor[HTML]{00441B}} \color[HTML]{F1F1F1} 0.405 & {\cellcolor[HTML]{EAF7E6}} \color[HTML]{000000} 0.056 & {\cellcolor[HTML]{EBF7E7}} \color[HTML]{000000} 0.053 & {\cellcolor[HTML]{F7FCF5}} \color[HTML]{000000} 0.021 & {\cellcolor[HTML]{005F26}} \color[HTML]{F1F1F1} 0.372 \\
& Avg. &  {\cellcolor[HTML]{5AB769}} \color[HTML]{F1F1F1} 0.289 & {\cellcolor[HTML]{58B668}} \color[HTML]{F1F1F1} 0.290 & {\cellcolor[HTML]{83CB82}} \color[HTML]{000000} 0.244 & {\cellcolor[HTML]{C9EAC2}} \color[HTML]{000000} 0.152 & {\cellcolor[HTML]{00441B}} \color[HTML]{F1F1F1} 0.475 & {\cellcolor[HTML]{F5FBF2}} \color[HTML]{000000} 0.055 & {\cellcolor[HTML]{ECF8E8}} \color[HTML]{000000} 0.081 & {\cellcolor[HTML]{F7FCF5}} \color[HTML]{000000} 0.047 & {\cellcolor[HTML]{00441B}} \color[HTML]{F1F1F1} 0.474 \\

\specialrule{.4pt}{0pt}{0pt}
\multirow{6}{*}{$\rho$} & Pop. & {\cellcolor[HTML]{084B93}} \color[HTML]{F1F1F1} 0.311 & {\cellcolor[HTML]{0A539E}} \color[HTML]{F1F1F1} 0.300 & {\cellcolor[HTML]{1C6AB0}} \color[HTML]{F1F1F1} 0.269 & {\cellcolor[HTML]{7AB6D9}} \color[HTML]{000000} 0.155 & {\cellcolor[HTML]{083B7C}} \color[HTML]{F1F1F1} 0.333 & {\cellcolor[HTML]{D3E3F3}} \color[HTML]{000000} 0.055 & {\cellcolor[HTML]{AED1E7}} \color[HTML]{000000} 0.107 & {\cellcolor[HTML]{F7FBFF}} \color[HTML]{000000} -0.012 & {\cellcolor[HTML]{08306B}} \color[HTML]{F1F1F1} 0.349 \\
& Inter. &  {\cellcolor[HTML]{08306B}} \color[HTML]{F1F1F1} 0.298 & {\cellcolor[HTML]{0E58A2}} \color[HTML]{F1F1F1} 0.261 & {\cellcolor[HTML]{9CC9E1}} \color[HTML]{000000} 0.150 & {\cellcolor[HTML]{7DB8DA}} \color[HTML]{000000} 0.168 & {\cellcolor[HTML]{08488E}} \color[HTML]{F1F1F1} 0.276 & {\cellcolor[HTML]{EAF3FB}} \color[HTML]{000000} 0.074 & {\cellcolor[HTML]{F7FBFF}} \color[HTML]{000000} 0.059 & {\cellcolor[HTML]{CDDFF1}} \color[HTML]{000000} 0.111 & {\cellcolor[HTML]{083573}} \color[HTML]{F1F1F1} 0.293 \\
& Comp. & {\cellcolor[HTML]{81BADB}} \color[HTML]{000000} 0.232 & {\cellcolor[HTML]{7DB8DA}} \color[HTML]{000000} 0.236 & {\cellcolor[HTML]{AACFE5}} \color[HTML]{000000} 0.178 & {\cellcolor[HTML]{B5D4E9}} \color[HTML]{000000} 0.163 & {\cellcolor[HTML]{08519C}} \color[HTML]{F1F1F1} 0.434 & {\cellcolor[HTML]{ECF4FB}} \color[HTML]{000000} 0.046 & {\cellcolor[HTML]{D9E7F5}} \color[HTML]{000000} 0.092 & {\cellcolor[HTML]{F7FBFF}} \color[HTML]{000000} 0.019 & {\cellcolor[HTML]{08306B}} \color[HTML]{F1F1F1} 0.494 \\
& Out. &  {\cellcolor[HTML]{0E59A2}} \color[HTML]{F1F1F1} 0.292 & {\cellcolor[HTML]{1B69AF}} \color[HTML]{F1F1F1} 0.272 & {\cellcolor[HTML]{1865AC}} \color[HTML]{F1F1F1} 0.278 & {\cellcolor[HTML]{8ABFDD}} \color[HTML]{000000} 0.165 & {\cellcolor[HTML]{08306B}} \color[HTML]{F1F1F1} 0.340 & {\cellcolor[HTML]{CEE0F2}} \color[HTML]{000000} 0.099 & {\cellcolor[HTML]{F7FBFF}} \color[HTML]{000000} 0.035 & {\cellcolor[HTML]{CFE1F2}} \color[HTML]{000000} 0.098 & {\cellcolor[HTML]{083674}} \color[HTML]{F1F1F1} 0.332 \\
& Evd.Inf & {\cellcolor[HTML]{4292C6}} \color[HTML]{F1F1F1} 0.337 & {\cellcolor[HTML]{5BA3D0}} \color[HTML]{F1F1F1} 0.300 & {\cellcolor[HTML]{4896C8}} \color[HTML]{F1F1F1} 0.329 & {\cellcolor[HTML]{DAE8F6}} \color[HTML]{000000} 0.099 & {\cellcolor[HTML]{08306B}} \color[HTML]{F1F1F1} 0.524 & {\cellcolor[HTML]{E4EFF9}} \color[HTML]{000000} 0.074 & {\cellcolor[HTML]{E5EFF9}} \color[HTML]{000000} 0.072 & {\cellcolor[HTML]{F7FBFF}} \color[HTML]{000000} 0.027 & {\cellcolor[HTML]{084285}} \color[HTML]{F1F1F1} 0.490 \\
& Avg. & {\cellcolor[HTML]{4A98C9}} \color[HTML]{F1F1F1} 0.406 & {\cellcolor[HTML]{4695C8}} \color[HTML]{F1F1F1} 0.412 & {\cellcolor[HTML]{6CAED6}} \color[HTML]{F1F1F1} 0.348 & {\cellcolor[HTML]{BFD8ED}} \color[HTML]{000000} 0.219 & {\cellcolor[HTML]{083674}} \color[HTML]{F1F1F1} 0.619 & {\cellcolor[HTML]{F2F8FD}} \color[HTML]{000000} 0.080 & {\cellcolor[HTML]{E6F0F9}} \color[HTML]{000000} 0.115 & {\cellcolor[HTML]{F7FBFF}} \color[HTML]{000000} 0.065 & {\cellcolor[HTML]{08306B}} \color[HTML]{F1F1F1} 0.633 \\
\specialrule{.4pt}{0pt}{0pt}
\multirow{6}{*}{acc\textsubscript{eq}} & Pop. & {\cellcolor[HTML]{760176}} \color[HTML]{F1F1F1} 0.459 & {\cellcolor[HTML]{840178}} \color[HTML]{F1F1F1} 0.454 & {\cellcolor[HTML]{A5017D}} \color[HTML]{F1F1F1} 0.444 & {\cellcolor[HTML]{F35F9F}} \color[HTML]{F1F1F1} 0.408 & {\cellcolor[HTML]{49006A}} \color[HTML]{F1F1F1} 0.476 & {\cellcolor[HTML]{FCBCBD}} \color[HTML]{000000} 0.374 & {\cellcolor[HTML]{FA99B3}} \color[HTML]{000000} 0.390 & {\cellcolor[HTML]{FFF7F3}} \color[HTML]{000000} 0.334 & {\cellcolor[HTML]{54006D}} \color[HTML]{F1F1F1} 0.472 \\
& Inter. & {\cellcolor[HTML]{CD238F}} \color[HTML]{F1F1F1} 0.332 & {\cellcolor[HTML]{F35F9F}} \color[HTML]{F1F1F1} 0.321 & {\cellcolor[HTML]{FDE1DE}} \color[HTML]{000000} 0.291 & {\cellcolor[HTML]{FDD9D6}} \color[HTML]{000000} 0.294 & {\cellcolor[HTML]{49006A}} \color[HTML]{F1F1F1} 0.357 & {\cellcolor[HTML]{FFF7F3}} \color[HTML]{000000} 0.282 & {\cellcolor[HTML]{FEECE9}} \color[HTML]{000000} 0.286 & {\cellcolor[HTML]{760176}} \color[HTML]{F1F1F1} 0.348 & {\cellcolor[HTML]{93017A}} \color[HTML]{F1F1F1} 0.343 \\
& Comp. & {\cellcolor[HTML]{FAA3B6}} \color[HTML]{000000} 0.378 & {\cellcolor[HTML]{FAA3B6}} \color[HTML]{000000} 0.379 & {\cellcolor[HTML]{FCC2BF}} \color[HTML]{000000} 0.364 & {\cellcolor[HTML]{FCCDC9}} \color[HTML]{000000} 0.357 & {\cellcolor[HTML]{5B006F}} \color[HTML]{F1F1F1} 0.461 & {\cellcolor[HTML]{FFF7F3}} \color[HTML]{000000} 0.328 & {\cellcolor[HTML]{FDE1DE}} \color[HTML]{000000} 0.345 & {\cellcolor[HTML]{FFF3EF}} \color[HTML]{000000} 0.331 & {\cellcolor[HTML]{49006A}} \color[HTML]{F1F1F1} 0.468 \\
& Out. & {\cellcolor[HTML]{EE579E}} \color[HTML]{F1F1F1} 0.337 & {\cellcolor[HTML]{F76CA3}} \color[HTML]{F1F1F1} 0.332 & {\cellcolor[HTML]{F76BA2}} \color[HTML]{F1F1F1} 0.333 & {\cellcolor[HTML]{FCD2CE}} \color[HTML]{000000} 0.302 & {\cellcolor[HTML]{49006A}} \color[HTML]{F1F1F1} 0.383 & {\cellcolor[HTML]{FDE6E2}} \color[HTML]{000000} 0.293 & {\cellcolor[HTML]{FFF7F3}} \color[HTML]{000000} 0.283 & {\cellcolor[HTML]{CD238F}} \color[HTML]{F1F1F1} 0.350 & {\cellcolor[HTML]{99017B}} \color[HTML]{F1F1F1} 0.363 \\
& Evd.Inf & {\cellcolor[HTML]{DF3898}} \color[HTML]{F1F1F1} 0.501 & {\cellcolor[HTML]{EA4F9C}} \color[HTML]{F1F1F1} 0.490 & {\cellcolor[HTML]{E13D99}} \color[HTML]{F1F1F1} 0.499 & {\cellcolor[HTML]{FCC1BF}} \color[HTML]{000000} 0.425 & {\cellcolor[HTML]{49006A}} \color[HTML]{F1F1F1} 0.586 & {\cellcolor[HTML]{FCCFCB}} \color[HTML]{000000} 0.412 & {\cellcolor[HTML]{FCD2CE}} \color[HTML]{000000} 0.408 & {\cellcolor[HTML]{FFF7F3}} \color[HTML]{000000} 0.367 & {\cellcolor[HTML]{7A0177}} \color[HTML]{F1F1F1} 0.558 \\
& Avg. & {\cellcolor[HTML]{B40881}} \color[HTML]{F1F1F1} 0.611 & {\cellcolor[HTML]{B30681}} \color[HTML]{F1F1F1} 0.611 & {\cellcolor[HTML]{D02690}} \color[HTML]{F1F1F1} 0.589 & {\cellcolor[HTML]{F767A1}} \color[HTML]{F1F1F1} 0.545 & {\cellcolor[HTML]{55006D}} \color[HTML]{F1F1F1} 0.676 & {\cellcolor[HTML]{FCBCBD}} \color[HTML]{000000} 0.482 & {\cellcolor[HTML]{FCBCBD}} \color[HTML]{000000} 0.481 & {\cellcolor[HTML]{FFF7F3}} \color[HTML]{000000} 0.402 & {\cellcolor[HTML]{49006A}} \color[HTML]{F1F1F1} 0.686 \\
\specialrule{.8pt}{0pt}{2pt}

\end{tabular}
\caption{Kendall's $\tau_{b}$, Spearman's $\rho$, and pairwise accuracy acc\textsubscript{eq} of systems to human evaluations. Note, each {\sc FactPICO} attribute is compared against the overall score produced by each system, which for LLM evaluators is the average rating across each element. We present an attribute-wise comparison for LLM evaluators in Appendix~\ref{sec:additional_corrtable}. }
\label{table:colorrow}
\end{table*}
\begin{table*}[t]
\centering
\small
\begin{tabular}{lrrrrrrrrr}
\toprule
{} &  \textbf{QAFact} & \textbf{Quest} & \textbf{AlignS} & \textbf{DAE} & \textbf{GPT-4} & \textbf{Llama-2} & \textbf{Alpaca} & \textbf{Mistral} & 
 \textbf{Extract}  \\
\midrule
ALPACA     &   \textbf{3.680} &  \textbf{0.547} &  \textbf{0.884} & \textbf{0.654} &  \textbf{3.608} &    \textbf{3.375} &   0.934 &    \textbf{3.619} &    
\textbf{3.277}\\
GPT-4      &   1.976 &  0.415 &  0.683 & 0.317 &  3.528 &    3.225 &   \textbf{0.942} &    3.297 &    2.891 \\
LLAMA-2    &   1.894 &  0.412 &  0.610 & 0.379 &  3.152 &    3.128 &   0.920 &    3.447 &    2.614 \\
\bottomrule
\end{tabular}
\caption{Average of systematic metrics per LLM for plain language summary. 
}
\label{table:grade}
\end{table*}
\section{Factuality Evaluation Metrics}%
{\sc FactPICO} is a %
dataset that can be used to assess automatic evaluation methods for plain language summarization of RCT texts. 
We assess existing methods used for factuality evaluation, as well as the capabilities of LLMs themselves to evaluate factuality. 
These analyses focus on questions about PICO elements and Evidence Inference; we leave evaluation of the factuality of added information for future work, as this entails fact-checking using external knowledge sources.

\subsection{Factuality Metrics Evaluated}\label{sec:factualitymodels} %

We first evaluate %
a suite of existing automatic factuality metrics shown to be effective in prior work. 
(1) \textbf{Dependency-Arc Entailment (DAE)} \cite{goyal2021annotating} decomposes summaries into smaller entailment tasks at the arc-level to assess their factuality. 
We use a numeric score by taking the minimum of the probability scores assigned to individual arcs.
(2) \textbf{QuestEval} \cite{scialom2021questeval} uses a QA-based framework to analyze the factual faithfulness of a summary to the original text. This method scores summaries a 0 if there is no common token and a 1 for an exact match.
(3) \textbf{QAFactEval} \cite{fabbri-etal-2022-qafacteval} is a QA-based metric that combines various components from other factuality metrics and assigns scores based on the \textit{LERC} score \cite{chen-etal-2020-mocha}. This score, usually used for evaluating reading comprehension answers, ranges from 1 to 5, where 1 is a completely wrong answer and 5 is a perfect answer.
(4) \textbf{AlignScore} \cite{zha-etal-2023-alignscore} is an alignment-based method for analyzing factual consistency. The final score assigned %
is the average of the maximum alignment probabilities between sentences from the summary to context chunks from the original abstract.

\subsection{LLM Evaluators}
\label{llm-eval}

Prior work has also shown that LLMs themselves can be good evaluators for factuality in summarization~\cite{luo2023chatgpt,wang2023chatgpt,tang-etal-2023-understanding,tian2023fine}. 
{\sc FactPICO} judgments are %
finer-grained. 
Therefore, we prompt %
LLMs with instructions emulating the questions asked of human evaluators in {\sc FactPICO}. 
In addition to %
ratings, the LLMs also generate rationales for their %
scores, which we analyze in Section~\ref{sec:prelim_rat}.

The implementation details and the prompts used for this task %
are in Appendix~\ref{sec:pico_prompts}. 
Prior work found that LLM evaluations may be biased in that a system may `prefer' its own outputs \cite{liu2023gpteval}, but we did not observe this here  (Table~\ref{table:grade}). 

\vspace{-0.5em}
\paragraph{Full-Text Evaluation.} %
As input, we provide %
LLMs %
the full-text of a complex medical abstract and corresponding LLM-generated plain language summary. 
We instruct the evaluator LLM to find PICO elements in the text %
and evaluate them according to the provided criteria.
In addition to PICO elements, we also evaluate Evidence Inference outputs; here the reference results span (annotated in prior work) is compared against the full text of the plain language summary. 
We evaluate four LLMs in this way: GPT-4~\cite{openai2023gpt4}, Llama-2-Chat (7B)~\cite{touvron2023llama}, Alpaca (7B)~\cite{alpaca}, and Mistral (7B-Instruct-v0.1; \citealt{jiang2023mistral}). 

\vspace{-0.35em}
\paragraph{PICO-R Extraction Pipeline.} %
We also evaluate LLM scorers explicitly informed of the PICO elements and results inferred from evidence (PICO-R).
We adopt this two-stage pipeline %
using GPT-4. %
We first extract PICO-R from both the original abstract and the plain language summary.\footnote{Initial experiments showed that GPT-4 can more accurately extract PICO elements compared to other LLMs and PICO-tagger models~\cite{nye-etal-2020-trialstreamer}, especially for plain language summaries.}
For Evidence Inference, this extraction is only necessary from the summary. %
Next, extracted elements are %
passed to GPT-4 along with an evaluation prompt. %

\section{Factuality Metric Evaluation Results}\label{sec:autoevalresults}

\begin{figure*}[t]
{\includegraphics[width=5.3cm, page=47]{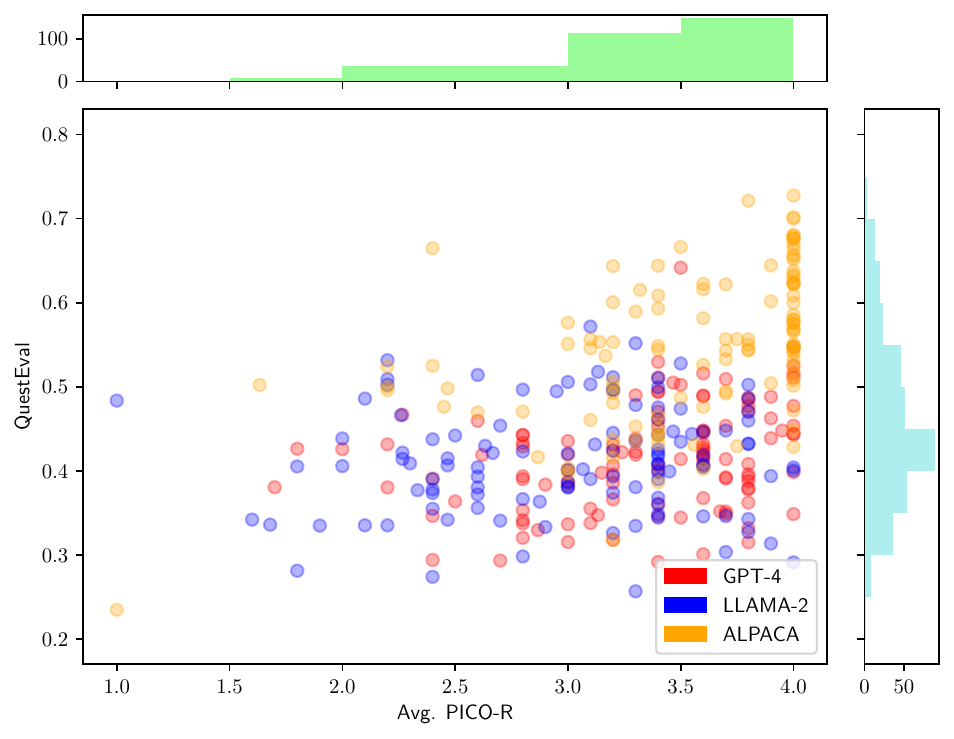}}
{\includegraphics[width=5.3cm, page=50]{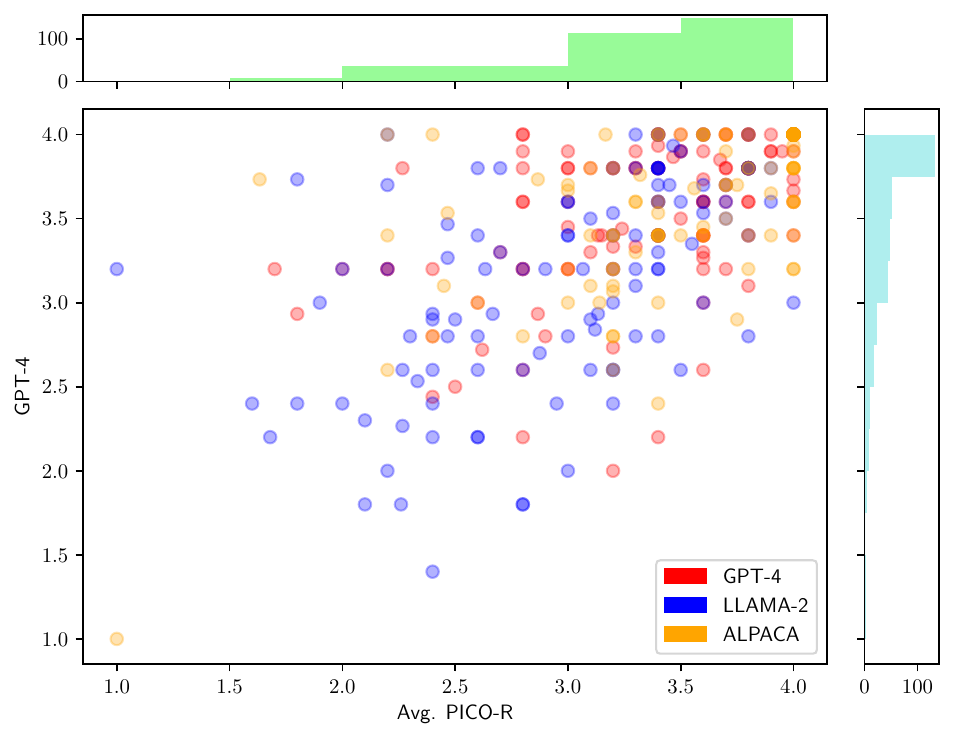}}
{\includegraphics[width=5.3cm, page=54]{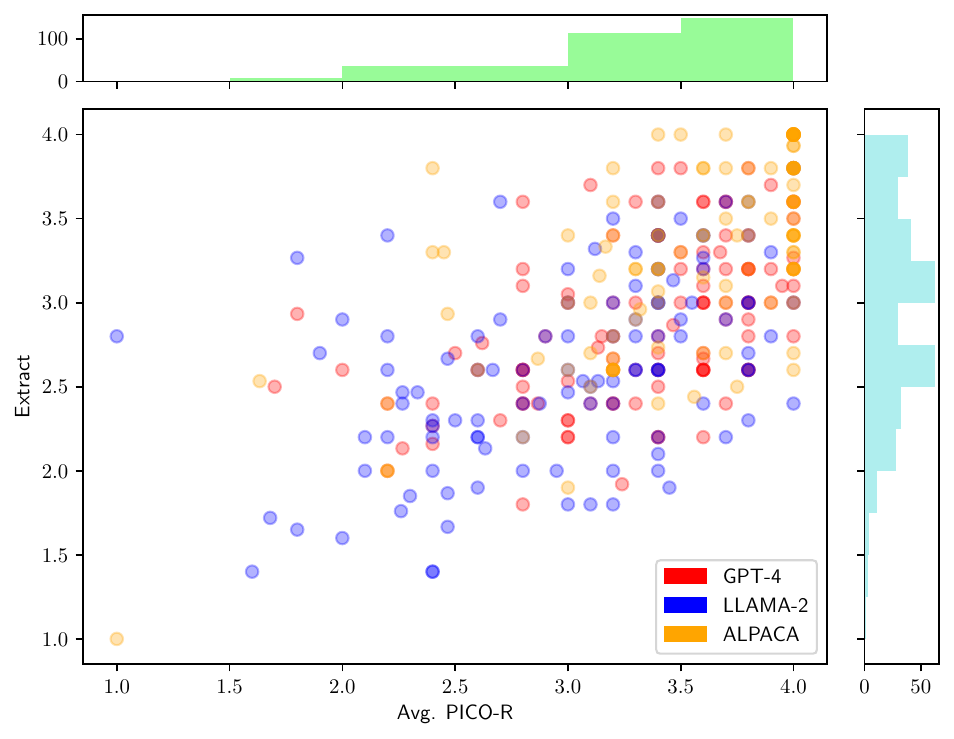}}
\caption{QuestEval (left), GPT-4 eval (mid), and Extract (right) against Avg. PICO-R (x-axis) for plain language summaries generated by GPT-4 (red), Llama-2 (blue), and Alpaca (orange). Label distributions shown on the sides.
}
\label{fig:scatt}
\end{figure*}

To conduct a meta evaluation of the system factuality metrics, we compute the Kendall's $\tau_b$ coefficient, Spearman correlation coefficient, and Pairwise Accuracy coefficient \cite{deutsch-etal-2023-ties} for automatic vs. human evaluations for each of the PICO and evidence inference aspect. 
We posit that good metrics must correlate well with the most salient elements in a high-stake domain.

\vspace{-0.5em}
\paragraph{Instance-level Results.} 
\label{sec:flipped}
We present the results for the above metrics in Table \ref{table:colorrow}.\footnote{The evaluation scale for Llama-2 was reversed because it struggled to follow the original instructions. We described this behavior in greater detail in Appendix~\ref{sec:llama2_analysis}}
Across most measures, with the exception of Kendall's $\tau_{b}$, the pipeline combining GPT-4 evaluation with PICO extraction has the highest correlations with the {\sc FactPICO} ratings. Thus, decomposing the original evaluation task into separate localization then evaluation steps  yield better performance, indicating that LLMs benefit from breaking apart a complex tasks into a series of simpler steps completed separately.

{\sc FactPICO} ratings correlated more with dedicated factuality models than LLM evaluators with the exception of GPT-4. In fact, the smaller, open-source LLMs had barely any correlation at all with the {\sc FactPICO} ratings. These results show that there is a sizeable gap in an LLMs ability to evaluate generated text compared to its generation capabilities.

\vspace{-0.5em}
\paragraph{System-level Results.} LLMs in Table \ref{table:grade}, with the exception of Alpaca, produce similar results compared to the human average PICO-R ratings in {\sc FactPICO} in Table \ref{table:llm_analysis}. %
Alpaca rates plain language summaries %
low, with averages under the minimum in the scale (1), and so has failed to follow the evaluation instructions. %
Although Alpaca-generated summaries have higher ratings, these aren't meaningful as are extractive.

\vspace{-0.5em}
\paragraph{Visual Analysis.} For more visual analysis of the system factuality metrics vs human ratings, Figure~\ref{fig:scatt} shows the top three performing systems vs. averaged human scores. The highest performing system, the PICO-R extraction pipeline, has a the most balanced linear pattern. On the other hand, GPT-4 without PICO-R extraction often rates non-factual summaries highly. Contrastively, QuestEval, the best performing non-LLM factuality metric, is more cautious and rates high quality summaries lower. We further analyze automatic vs.\ human rating distributions in Appendix~\ref{sec:more_scatters}. 
These results hint at the potential challenges of factuality assessment brought by the shift in readability.

\section{Preliminary LLM Rationale Analysis}
\label{sec:prelim_rat}
We perform a preliminary analysis on LLM rationales, comparing to expert rationales in {\sc FactPICO}; we leave a thorough human evaluation of LLM rationales for future work. 

Qualitative analysis of a small sample of rationales show that Llama-2 and Mistral are often able to comprehend the medical text but fail in making correct judgments according to the provided instructions. Most commonly, Mistral 
focuses on the abstract rather than the evaluated summary, while
Llama-2
generates long explanations that eventually arrive at the wrong conclusion.
Rationales from GPT-4 and its pipelined counterpart usually did not do this. %
For the most part, rationales made logical sense. 
However, %
some rationales were overly generous in it evaluation, %
ignoring critical errors. 
We provide examples of such erroneous rationales in Appendix~\ref{sec:rat_errors}.

\begin{table}[t]
\centering
\small
\resizebox{\linewidth}{!}{
\begin{tabular}{lrrrrr}
\toprule
{} &  \textbf{GPT-4} & \textbf{Llama-2} & \textbf{Alpaca} & \textbf{Mistral} & \textbf{Extract} \\
\midrule
\textbf{P}     &   0.189 &  0.073 &  0.153 \textsubscript{0.21} & 0.156 &  \textbf{0.201} \\
\textbf{I}     &   0.173 &  0.073 &  \textbf{0.210} \textsubscript{0.11} & 0.169 &  \textbf{0.205} \\
\textbf{C}    &   0.181 &  0.092 &  0.126 \textsubscript{0.11} & 0.173 &  \textbf{0.215}  \\
\textbf{O}   &   0.148 &  0.075 &  0.111 \textsubscript{0.19} & 0.158 &  \textbf{0.176}  \\
\textbf{R}   &   \textbf{0.061} &  0.019 &  0.045 \textsubscript{0.60} & 0.045 &  0.060  \\
\bottomrule
\end{tabular}}
\caption{BERTScore (rescore baseline) F1 average for PICO and Evidence Inference (R). Note, the results for Alpaca omit rationales. We show the percentage of non-empty rationales next to the BERTScore for Alpaca. We caution comparing the results for Alpaca here against those of other LLMs because instances of empty rationales have been excluded in this evaluation.}
\label{table:bertscores}
\end{table}

Table~\ref{table:bertscores} shows the BERTScore~\cite{zhang2020bertscore}
between expert vs.\ LLM rationales across all evaluators experimented in this work.
Overall, the rationales are dissimilar.
The PICO-R extraction pipeline produced rationales most similar to that of humans, with the exception of rationales for evidence inferences, where GPT-4 rationales are slightly more similar than others. 
Despite explicit prompting, Alpaca frequently did not produce rationales and stopped generation after outputting a numerical rating.

Despite reporting these preliminary  results, we advise caution when adopting reference-based metrics for automatic free-text rationale evaluation, as the results may be misleading.  
For example, we observed that Alpaca rationales tend to be a somewhat arbitrary variation of the rating descriptions (e.g., \textit{The interventions were described accurately}), and human evaluators often use the words in these descriptions in their rationales (e.g.,  \textit{The intervention is described correctly as an 8 week program of specific exercises}), inflating word overlap scores. %
By contrast, an example Llama-2 rationale (\textit{The comparator in the PICO model is the placebo group...}) could present the same idea as a human rationale (ex: \textit{Comparator (placebo) is mentioned.}), but %
result in a low BERTScore (ex: \textit{-0.182}) because rationales from Llama-2 tend to (much) longer than those written by humans. 
Future work developing metrics for rationale correctness should not solely rely on reference-based metrics. 
We additionally present a length analysis in Appendix~\ref{sec:rat_length}.%

\section{Related Work}

Meta-evaluation of factual consistency metrics in summarization (and related tasks) in the ``general domain'', e.g., Wikipedia and news, has garnered considerable attention \cite{pagnoni2021understanding, honovich-etal-2022-true-evaluating, laban-etal-2022-summac,tang-etal-2023-understanding,min2023factscore}. 
However, in addition to focusing on  different domains, these existing benchmarks in summarization include primarily older models. 
Newer LLMs may yield novel error types (or be more factual overall) ~\cite{tang-etal-2023-understanding}. 

Existing summarization factuality benchmarks also fail to generalize to {\em simplification}, in which content addition in the form of elaborations or explanations is often necessary~\cite{srikanth-li-2021-elaborative}. 
Broadly, simplification entails substantial language changes that often lead to the text being more general~\cite{speciteller}. 
\citet{devaraj-etal-2022-evaluating} evaluated the factuality of automated simplification model outputs at the sentence-level, noticing that content {\em deletion} can often lead to factual errors (in contrast to only summarization). Our findings confirm that these errors also exist in {\em plain language} summarization; the overgeneralization problem may lead to safety issues in the medical domain.

\citet{shaib-etal-2023-summarizing} and \citet{tang2023evaluating} evaluated LLM-generated summaries of medical evidence. Notably, \citet{shaib-etal-2023-summarizing}'s work included an evaluation of {\em plain language} summaries.
Our work deepens this analysis with a finer-grained evaluation focusing on critical components of RCTs and medical evidence, covering three LLMs. 
Our findings call for caution against LLM-generated plain language summaries despite the absence of outright inconsistencies. \citet{pal2023med} presents an analysis of ``hallucinations'' in medical QA tests, focusing on reasoning rather than factual consistency.

The inclusion of human-written natural language rationales in factuality benchmarks is rare, %
and there is a paucity of work evaluating these. 
The FELM benchmark \cite{chen2023felm}, an open-domain evaluation of LLM-generated long form texts covering factual knowledge, math, and reasoning included human rationales. 
Work in LLM critiquing has  started to incorporate natural language critiques from both humans~\cite{saunders2022self} and LLMs~\cite{cui2023ultrafeedback,kim2024prometheus}.
{\sc FactPICO} is the first factuality benchmark of the plain language summarization task that includes expert-generated natural language rationales.

\section{Conclusions} %

We introduced {\sc FactPICO}, an expert-annotated benchmark in the domain of evidence-based medicine for evaluating the factuality of plain language summarization with respect to clinically important dimensions. 
Using {\sc FactPICO}, we presented an analysis of factual errors along these fine-grained aspects in LLM-generated plain language summaries. 
We also presented an analysis of methods to evaluate factuality, including both dedicated factuality models and novel LLM-based methods.

\section*{Limitations} %

The process of evaluating texts in {\sc FactPICO} was time consuming for human evaluators, requiring close reading of complex, technical language. Consequently, to %
make the workload manageable, we did not ask evaluators to localize PICO elements (or results regarding these) in texts in plain language summaries. 
Annotating how these elements are represented overall in the summary would provide more insights in cases where they are vaguely represented and thus required more focus in this evaluation. 
We automatically localize these elements in the evaluated PICO-R extraction pipeline, but we encourage future work expanding {\sc FactPICO} to include human-annotated, span-level annotations.

We primarily evaluated the zero-shot capabilities of LLMs in evaluating the factuality of simplified medical texts. 
We chose this setting as it best reflects %
how an end-user---a lay individual---would likely interact with an LLM, as such users are unlikely to provide LLMs with expert-evaluated plain language summaries for few-shot prompting. 
Future work could explore other LLM evaluation methods methods and use {\sc FactPICO} as an evaluation benchmark.

In our meta-evaluation we compared factuality metrics that assess the overall factuality of a text against numerical ratings assessing the fine-grained factuality of key characteristics in these texts. We acknowledge that this is not an equivalent comparison. 
However, we posit that the aggregation of these assessments should be well-correlated with the overall factuality of RCT texts for them to be useful in this important domain.

\section*{Acknowledgments}
We thank David Heineman for help with
the annotation interface, Yao Dou for help with Alpaca, and Tanya Goyal and Greg Durrett for useful discussions. This research is partially supported by NSF CAREER Award IIS-2145479 and Good Systems,\footnote{\url{https://goodsystems.utexas.edu}} a UT Austin Grand Challenge to develop responsible AI technologies.
Trienes, Göke and Coers were supported by the Cancer Research Center Cologne Essen (CCCE). Trienes was also supported by the Federal Ministry of Education and Research (BMBF) and by a fellowship within the IFI programme of the German Academic Exchange Service (DAAD).
Wallace was supported in this work by the National Institutes of Health (NIH), grant R01LM012086, and by the National Science Foundation (NSF), grant 1750978. Xu is supported in part by NSF awards IIS-2144493 and IIS-2112633.

\bibliography{anthology,custom}

\clearpage
\appendix

\section{Evaluation Guidelines}
\begin{table*}[t]
\centering
\small
\begin{tabular}{llll}
\toprule
\textbf{PICO Element} &  \textbf{Description} &  \textbf{Critical Descriptors} &  \textbf{Example(s)} \\
\midrule
\textbf{P}opulation     & The types of patients &   \tabitem Demographics &  \tabitem Diabetic males\\
&  involved in the trial & \tabitem Specific condition & \tabitem Healthy adults \\
\midrule
\textbf{I}ntervention      & The treatments considered &   \tabitem Mentioned as an intervention &  \tabitem Aspirin\\
& & \tabitem Differentiates from comparator & \\
\midrule
\textbf{C}omparator      & The alternative treatment &   \tabitem Mentioned as comparing &  \tabitem Placebo \\
& to which the intervention & \: \: against intervention & \\
& is being compared against. & \tabitem Differentiates from intervention & \\
\midrule
\textbf{O}utcome      & The measures used. NOT & \tabitem Primary outcomes mentioned   & \tabitem Mortality \\
& what was found in the study& \: \: (cannot make any conclusions & \tabitem Duration of \\
& ("result"). For example, if the & \: \: without them). & \: \: headache \\
& study finds a drug reduces the & & \\
& duration of headache, the & & \\
& outcome here is just the  & & \\
& "duration of headache", & & \\
& not that it reduced it. & & \\
\bottomrule
\end{tabular}
\caption{PICO elements. A critical descriptor is a characteristic that is absolutely crucial to understanding the study.}
\label{table:critcial}
\end{table*}
\label{sec:eval_guide}
Annotators are shown a ``source” input (abstract of a technical paper describing an RCT), along with a plain language summary of it, automatically produced by a model, and asked a series of 7 questions for evaluation. \\ \\
\textbf{Question I: Added Information} \\ For this part of the question, please highlight words, phrases, or sentences in the output that adds or modifies the information from the original document. Afterwards, answer whether this added/modified information is factual or not factual. Also, provide a rationale for why this is added/modified information and why it is factual or not factual. \\ \\
\textbf{Questions II-V (one analogous question for each PICO element):}\\
For instance, for \textbf{population}, we ask:
\begin{itemize}
    \item[$\square$]The population is mentioned, and described accurately
    \item[$\square$]The population is mentioned, but described somewhat inaccurately or vaguely
    \item[$\square$]The population is mentioned, but described with severe inaccuracies and/or is missing critical descriptors
    \item[$\square$]The population is missing in the model summary
    \item[$\square$]N/A
\end{itemize}
Please provide a rationale for why you chose the answer choice in relation to the abstract and the summary. 

Some examples will actually not quite be randomized trials (e.g., they might be observational studies, or a description of a prospective trial not yet run). In these cases, it may be that there is no meaningful population (or intervention, comparator, and outcome). Here you should select the “N/A” option. 

\textbf{Exhaustive Outcomes} \emph{(Analyzed in Appendix~\ref{sec:exhaustive})}: Sometimes, the plain language summary may not mention all the outcome measures described in the abstract. The summary may still be considered factual if the omitted measures are non-critical for the experiment and are not mentioned any further in the abstract. However, please do annotate separately when the plain language summary does exhaustively mention all outcome measures and when it does not. \\

\noindent
\textbf{Question VI: Evidence Inferences}\\
You will be presented with a span of text highlighting the inferred result from the experiment presented in the abstract. Based on this span, choose the following based on how this span is presented in the plain language summary:

\begin{itemize}
    \item[$\square$] Accurate
    \item[$\square$] Vague/Slightly Inaccurate
    \item[$\square$] Inaccurate
    \item[$\square$] Not mentioned
\end{itemize}
Please provide a rationale for why you chose the answer choice in relation to the evidence inference span and the summary.

\subsection{Additional Questions}
We also collected information for any additional comments on the generated plain language summary, as well as contradictions that are {\em not} covered by the other questions. These annotations are scarce, thus we have not included them in the {\sc FactPICO} benchmark.

\textbf{Additional comments:} ``asks you to write down any commands you would want the machine to follow if you could interact with it, e.g., “Make it shorter”, “Explain XXX a bit more”, and so on.''

\textbf{Contradictions:}
``Here you will be looking for content in the output that contradicts some part of the input. We ask you to annotate both the input and the output for this question. Please provide a rationale for why the content is a contradiction.''
The contradictions are analyzed in Appendix~\ref{app:contra}.

\noindent 

\section{Data Release and License}

We reused RCT abstracts from the \emph{Evidence Inference V2.0} dataset~\citep{deyoung2020evidence}; {\small \url{https://evidence-inference.ebm-nlp.com}}, accessed 2024-02-15). All articles in this dataset are from the PubMed Open Access subset which only includes license terms that allow reuse ({\small \url{https://ncbi.nlm.nih.gov/pmc/tools/openftlist}}, accessed 2024-02-15). After discussion with our institutions' librarian on fair use, we release the annotations in {\sc FactPICO} under CC-BY-4.0.

\section{Improving Agreement Through Discussion}
\begin{table}[t]
\small 
    \centering
    \begin{tabular}{lc}
    \toprule
    \textbf{Type} & \textbf{$\kappa$}\\  \midrule
        Population & 0.47/0.56  \\
        Intervention & 0.59/0.80  \\
        Comparator & 0.63/0.73 \\
        Outcome & 0.56/0.60 \\
        \bottomrule
    \end{tabular}
    \caption{Inter-evaluator agreement measured through Randolph's $\kappa$;
    for PICO evaluations we show agreement on all 75 doubly annotated documents (left) and only the subset of 15 undiscussed documents (right).
    }
    \label{table:agg2}
\end{table}
\label{sec:discussion}

Initially, we had observed low agreement on PICO questions among evaluators on the first 30 doubly annotated documents. Upon deeper analysis of these disagreements, we modified our instructions to be clearer and more detailed. We had asked evaluators to independently re-evaluate the previous annotations as well as doubly annotate 30 more documents. We observed that while agreement had improved significantly overall, evaluators still disagreed substantially on questions regarding population and outcome. 

To fix this issue, we facilitated a ``soft" discussion between evaluators regarding their annotations. Evaluators were presented with documents in which they had disagreed majorly on any PICO questions from the last 30 documents they evaluated. Then they were asked to come to consensus on how to rate these questions. Afterwards, we asked evaluators to independently reevaluate their previous annotations based on the insights they gained from this discussion. Similarly, they were asked to doubly annotate 15 new documents independently. 

Table~\ref{table:agg2} shows the resulting agreement through Randolph's kappa after this discussion. For each question type, the first number is the kappa for the entire set of 75 doubly annotated documents. The second number
is the kappa for the set of 15 documents that were annotated independently after the discussion. Both sets of number show moderate to high agreement for these questions. 
Furthermore, the agreement for this undiscussed set being significantly higher than the agreement for all 75 doubly annotated documents indicates that this discussion method was effective at improving agreement.

\section{Model Details and Compute}
We used a High-RAM T4 GPU through Google Colab Pro+ to conduct our experiments. 
\subsection{Plain Language Summary Generation}

\textbf{GPT-4} We used a frequency penalty of 0, presence penalty of 0, temperature of 1, and top p of 1. \\
\textbf{Llama-2 7B Chat}. We set the max new tokens to 4000, did multinomial sampling, temperature of 1, top k of 50, and top p of 1.0.\\
\textbf{Alpaca (7B)}. We set the max new tokens to 4000, used greedy decoding, temperature of 1, top k of 50, and top p of 1.0.

\subsection{LLM evaluation}
\label{sec:llm_implem_details}

\paragraph{GPT-4} We used a frequency penalty of 0, presence penalty of 0, temperature of 1, and top p of 1.

\paragraph{Together.AI} We ran experiments with \texttt{Llama-2 7B Chat}, \texttt{Alpaca (7B)}, and \texttt{Mistral-7B-Instruct-v0.1} using the \href{https://www.together.ai/}{Together.AI} API interface. For all models, we set the max new tokens to 256, temperature of 0.6, top k of 90, top p of 0.8, and a repetition penalty of 1.1. 

\section{Zero-shot Prompts for Plain Language Summarization/Simplification}
\label{sec:prompts}
\subsection{Preliminary Prompt Exploration and Model Evaluation}
We conducted a preliminary evaluation in Table \ref{table:prem} for prompt engineering and model exploration on 100 medical abstracts from \citet{shaib-etal-2023-summarizing}. The prompts we tested for are all \textit{typical-use} prompts, representing how these systems would be usually used by the public. We avoided engineering prompts for the best-case use of these systems for this task, as we did not want to induce a false trust in any of these systems for medical use, which could be potentially harmful. Given this criteria, we evaluated the following prompts:\\ \\
\textbf{Paper Plain:} \cite{august2023paperplain}

\textit{My fifth grader asked me what this passage means: [abstract] I rephrased it for him, in plain language a fifth grader can understand:}\\ \\
\textbf{Short}: 

\textit{"My fifth grader asked me what this passage means: [abstract] Help me summarize it for him, in plain language a fifth grader can understand. Make it short."}\\ \\
\textbf{Summarize:}

\textit{"My fifth grader asked me what this passage means: [abstract] Help me summarize it for him, in plain language a fifth grader can understand."} \\ \\
\textbf{5th grade:}

\textit{5th Grade: Paraphrase this passage completely in your own words. Always define words the reader may not know: [abstract]}\\ \\
\textbf{Complex:} 

\textit{"Below is an instruction that describes a task, paired with an input that provides further context. Write a response that appropriately completes the request.\\ \\\#\#\# Instruction:\\Rewrite the following complex passage in order to make it easier to understand by non-native speakers of English.\\ \\\#\#\# Input:"}\\
\textit{[abstract] \\ \\\#\#\#Response:"} \\ \\
\textbf{Medical:} 

\textit{"Below is an instruction that describes a task, paired with an input that provides further context. Write a response that appropriately completes the request.\\ \\\#\#\# Instruction:\\Rewrite the following medical abstract in order to make it easier to understand by non-native speakers of English.\\ \\\#\#\# Input:"}\\
\textit{[abstract] \\ \\\#\#\#Response:"}
\subsection{{\sc FACTPICO} Prompts}
For GPT-4, we randomly sampled from GPT-4\textsubscript{Summarize} and GPT-4\textsubscript{Short}. We also used Llama-2\textsubscript{Paper Plain} and Alpaca\textsubscript{Complex}.
\begin{table}
\centering
\small
\begin{tabular}{lrrr}
\toprule
 \textbf{Model\textsubscript{Prompt}} &  \textbf{DA} & \textbf{FK} & \textbf{\#tokens} \\
\midrule
GPT-4\textsubscript{Paper Plain} & 85.93 & 9.155 & 216.77 \\
GPT-4\textsubscript{5th grade} & 89.90 & 10.606 & 308.73 \\
\textcolor{Green}{GPT-4\textsubscript{Summarize}} & 84.90 & 9.583 & 183.71 \\
\textcolor{Green}{GPT-4\textsubscript{Short}} & 85.21 & 14.741 & 111.28 \\
Flan-T5\textsubscript{Plain} & 87.53 & 14.741 & 47.43 \\
Flan-T5\textsubscript{5th grade} & 81.44 & 15.031 & 28.24 \\
\textcolor{Green}{LLAMA-2\textsubscript{Paper Plain}} & 81.03 & 8.218 & 135.38 \\
\textcolor{Green}{ALPACA\textsubscript{Complex}} & 88.41 & 13.308 & 113.21 \\
ALPACA\textsubscript{Medical} & 89.28 & 13.523 & 101.37 \\
Dataset & - & 11.879 & 293.74 \\

\bottomrule
\end{tabular}
\caption{ChatGPT-DA \cite{wang2023chatgpt}, Flesch-Kincaid Grade Level, and \# of tokens with spaCy tokenizer for preliminary evaluation. The Model\textsubscript{Prompt} we chose for {\sc FactPICO} simplifications are in \textcolor{Green}{green}.}
\label{table:prem}
\end{table}

\subsection{Explanation of Selection Criteria}
We primarily selected prompts based on the length of the plain language summaries produced when they are used. They had to be substantially shorter than their corresponding abstracts, qualifying them as plain language \textit{summaries}. We did not use any form of length control outside of prompt instructions, as we found prompt instructions, such as "Make it short.", are just as effective as other forms of length control without creating any unwanted artifacts and disfluencies in the text. We did not consider ChatGPT-DA~\cite{wang2023chatgpt} to be a reliable measure of factuality. However, it was an effective sanity check in determining the relevance of the summaries to their corresponding abstracts. 

We included two prompts to get generations from GPT-4 because both prompts met the above selection criteria and added more diversity in {\sc FactPICO}. As a benchmark for factuality evaluation methods, including more diverse outputs will ensure only very robust systems are capable of attaining good results on {\sc FactPICO}.

\section{Analysis of PICO Specific LLM Evaluations}
\label{sec:additional_corrtable}
\begin{table*}
\centering
\small
\begin{tabular}{lrrrrrrrrrr}
\toprule
 &  & \textbf{GPT-4} &  \textbf{Llama-2} &  \textbf{Alpaca} &  \textbf{Mistral} &  \textbf{Extract} \\
\midrule
\multirow{6}{*}{$\tau_{b}$} & Pop.  &  0.323 &  0.017 &   0.095 &-0.052 &    \textbf{0.377} \\
& Inter. &  \textbf{0.233} &            0.053 &   0.036 &0.081 &    0.222 \\
& Comp.    &  0.506 &            0.001 &   0.064 &    -0.001 &    \textbf{0.575} \\
& Out.     &  0.260 &            0.086 &   0.090 &    0.145 &    \textbf{0.261} \\
& Evd. Inf.     &\textbf{0.561}&            0.137 &   -0.045 &    0.068 &    0.506 \\
& Avg.    &  \textbf{0.475} &            0.055 &   0.081 &    0.047 &    0.474 \\
\midrule
\multirow{6}{*}{$\rho$} & Pop.     &0.363 &            0.019&   0.110 &   -0.059 &    \textbf{0.429} \\
& Inter.   & \textbf{0.246}&            0.058 &   0.040 &    0.089 &\textbf{0.246} \\
& Comp.     &  0.558 &            0.001 &   0.072 &    -0.001&    \textbf{0.649}\\
& Out.        &  0.279 &            0.094 &   0.100 &    0.159 &    \textbf{0.292}\\
& Evd. Inf.     &\textbf{0.66}7 &            0.170 &  -0.051 &0.078 &    0.604 \\
& Avg.      &  0.619 &            0.080 &   0.115 &    0.065 &    \textbf{0.633} \\
\midrule
\multirow{6}{*}{acc\textsubscript{eq}}   & Pop.     &  0.469 &            0.331 &   0.365 &   0.287 &    \textbf{0.482} \\
& Inter.   & \textbf{0.538}&            0.379 &   0.365 &    0.420 &    0.419 \\
& Comp.     &  0.604 &            0.339 &   0.366 &    0.343 &    \textbf{0.632}\\
& Out.        &  \textbf{0.502} &            0.381 &   0.383 &    0.450 &    0.423\\
& Evd. Inf.     &  \textbf{0.665} &            0.419 &  0.342 &    0.287 &    0.631 \\
& Avg.    &  0.676 &            0.482 &   0.481 &    0.402 &    \textbf{0.686} \\
\bottomrule
\end{tabular}
\caption{Kendall's $\tau_{b}$, Spearman's $\rho$, and pairwise accuracy acc\textsubscript{eq} of PICO specific systems to human evaluations. }
\label{table:picospecific}
\end{table*}

We present the attribute-wise results for each LLM-based system we evaluated in Table~\ref{table:picospecific}. Similar to the results presented in Table~\ref{table:colorrow}, the GPT-4 based systems are more correlated with human ratings than the other LLM evaluators. Interestingly, we do see that for some attributes, in particular \emph{Population} and \emph{Evidence Inference}, the full text GPT-4 approach produced more human correlated ratings compared to those produced using the extraction pipeline. This suggests that evaluation of these elements requires more context than that provided in the extracted spans to perform a more accurate evaluation. However, despite this result, in terms of overall factuality, the extraction pipeline is slightly more correlated than the full text approach, suggesting an overall benefit to decomposing a complex task such as this into smaller, simpler steps to be executed sequentially.

\section{System to Avg. PICO-R Visualizations}
\label{sec:more_scatters}

\begin{table}
\small 
    \centering
    \begin{tabular}{lllll}
    \toprule
    \textbf{Method} & \textbf{KL} \\  \midrule
        GPT-4 &  0.016\\
        GPT-4 + PICO Extract & 0.111 \\
        LLAMA-2 & 0.161 \\
        ALPACA & 0.192 \\
        Mistral & 0.042 \\
        DAE & 0.238 \\
        AlignScore & \textbf{0.015} \\
        QAFactEval & 0.283 \\
        QuestEval & 0.406 \\
        \bottomrule
    \end{tabular}
    \caption{KL divergence between the standardized distributions of evaluated metrics and that of the averaged PICO rating in {\sc FactPICO}}
    \label{table:kldiv}
\end{table}
\begin{figure}[t]
    \centering
    \includegraphics[width=\linewidth]{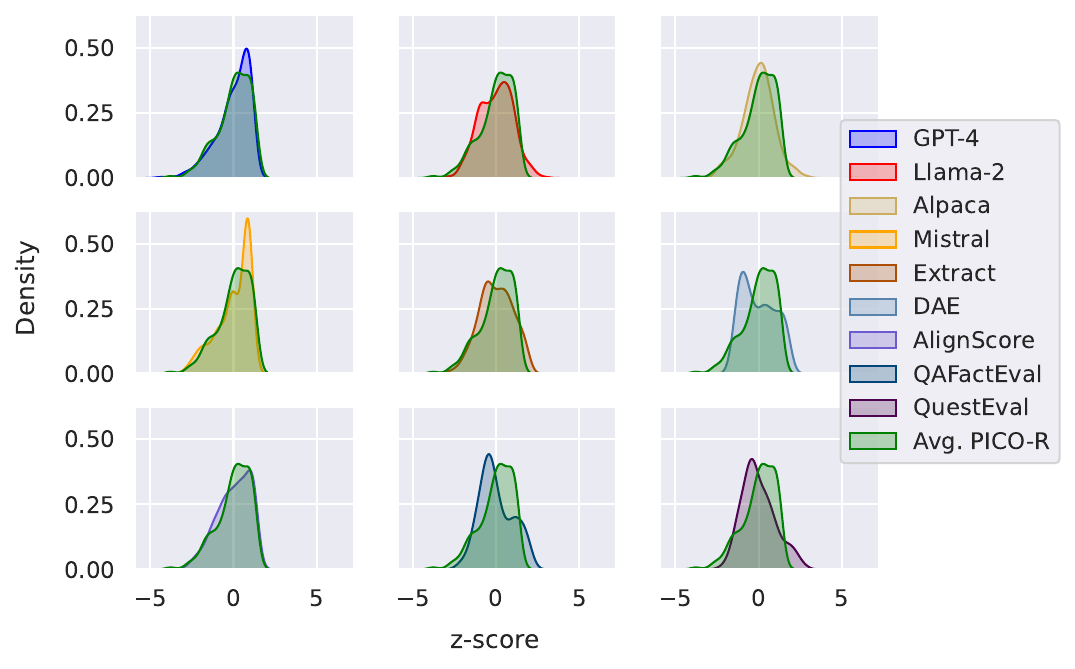}     
    \caption{Plots of estimated Gaussian probability density functions from the standardized distributions of evaluated metrics.}
    \label{fig:dist_gauss}
\end{figure}

\begin{figure*}
{\includegraphics[width=7.5cm, page=46]{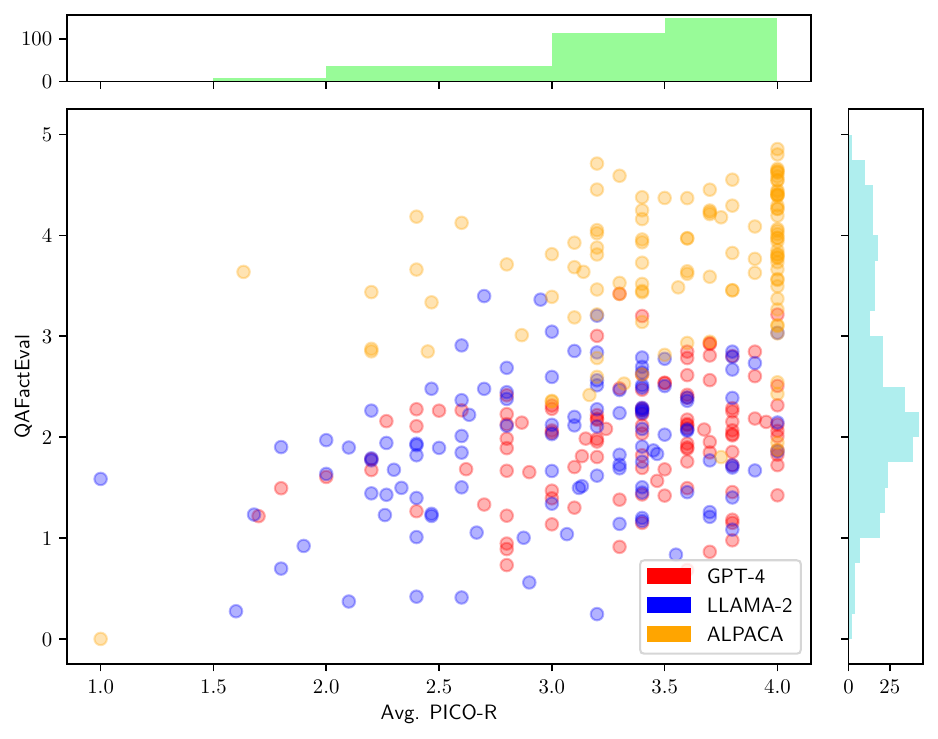}}
{\includegraphics[width=7.5cm, page=47]{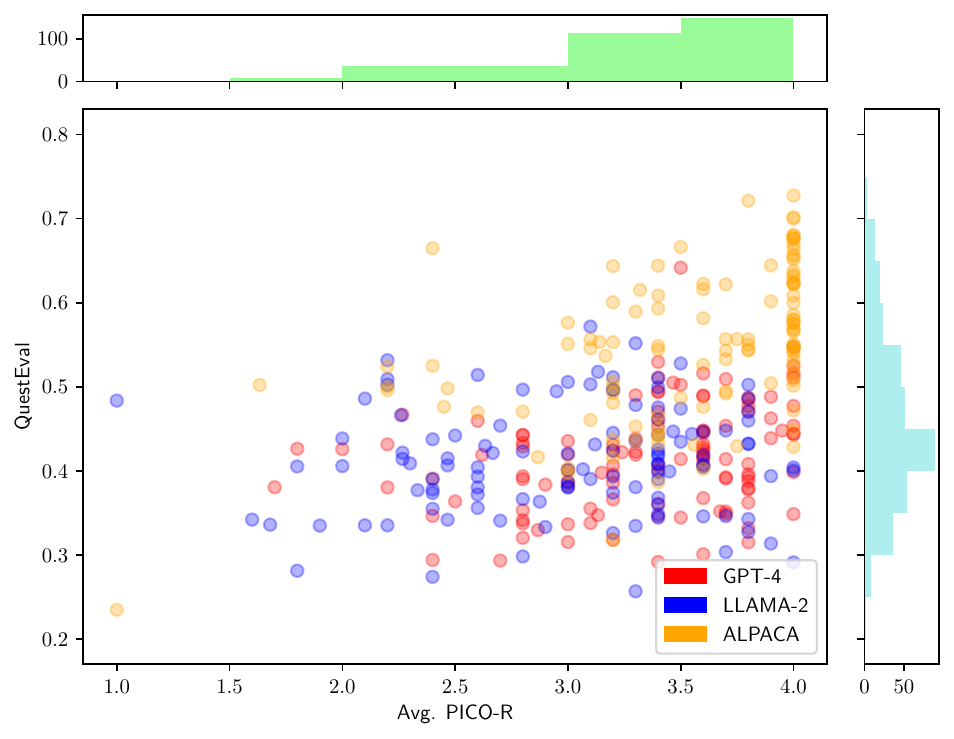}}\\
{\includegraphics[width=7.5cm, page=48]{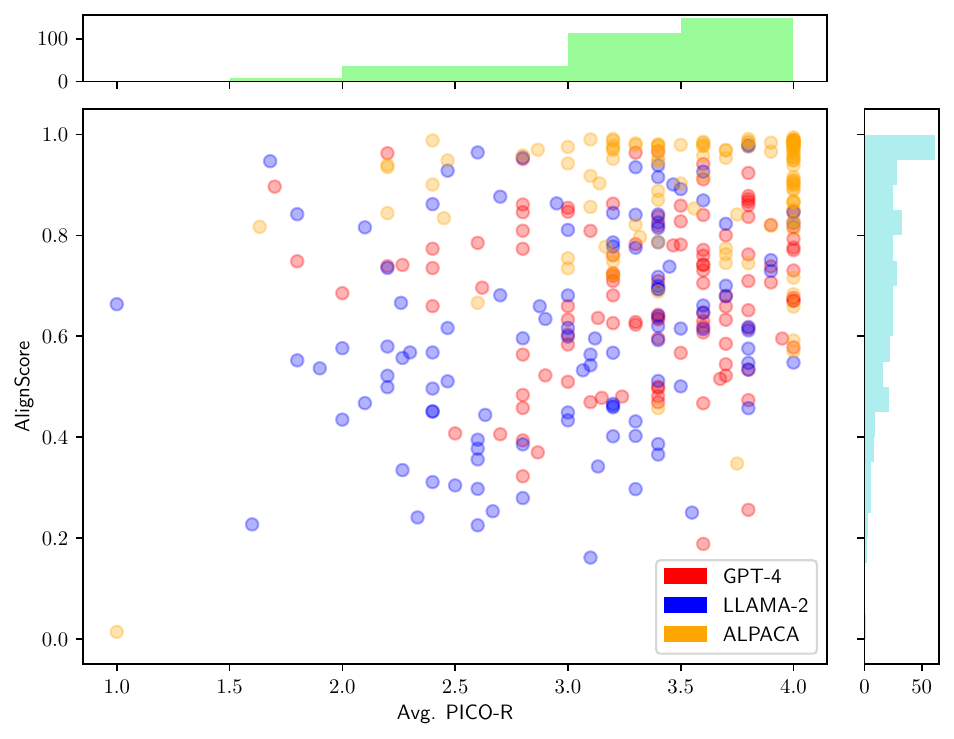}}
{\includegraphics[width=7.5cm, page=49]{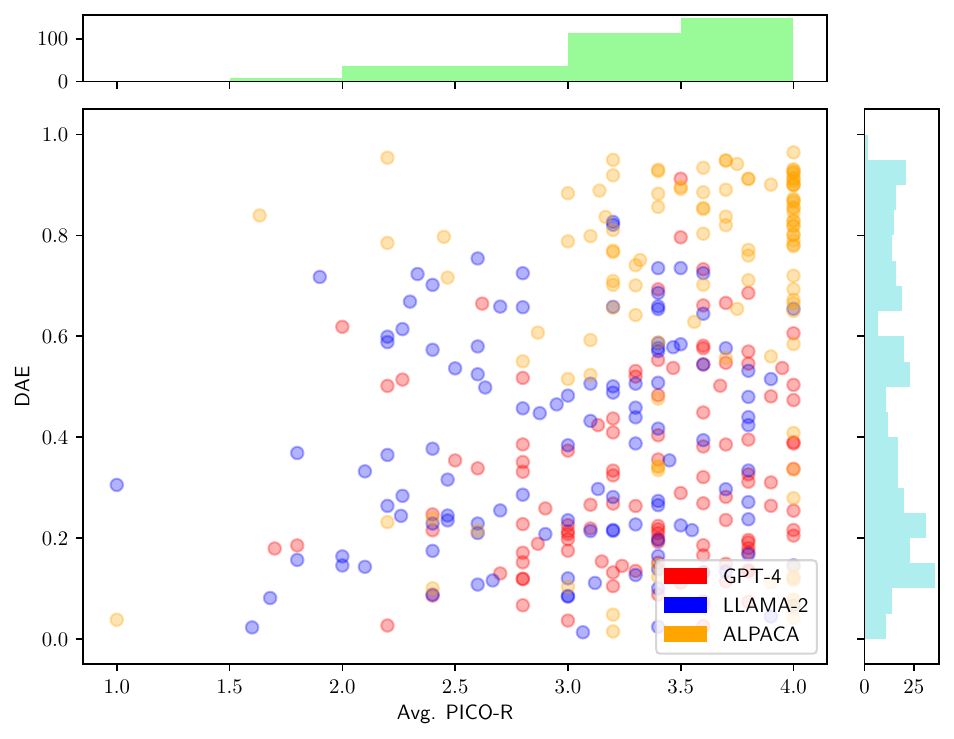}}\\
{\includegraphics[width=7.5cm, page=50]{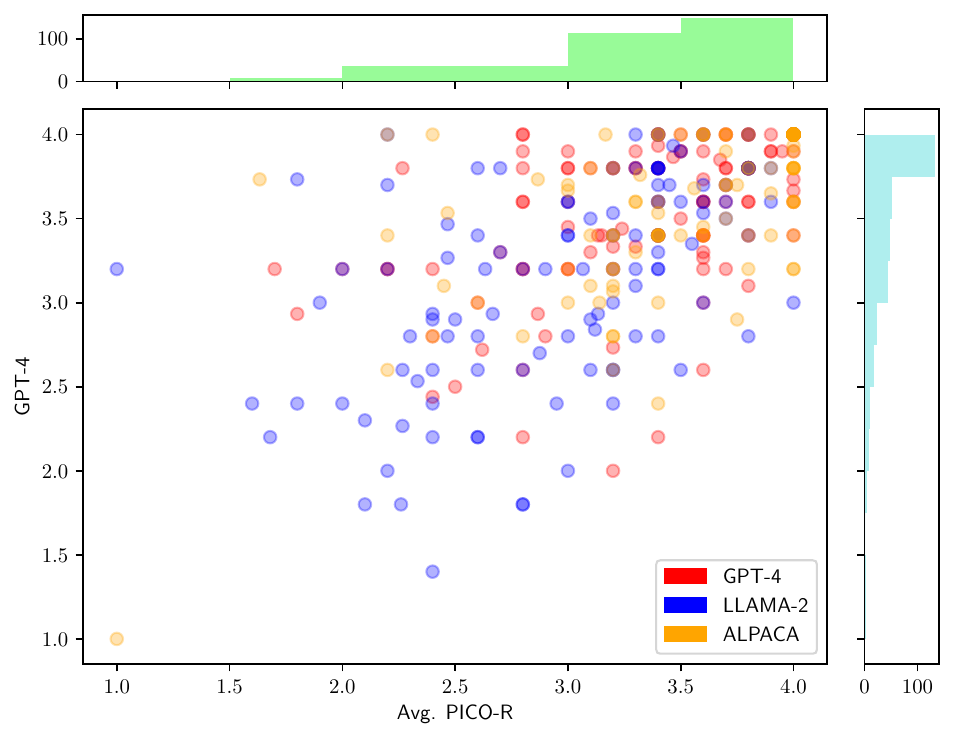}}
{\includegraphics[width=7.5cm, page=51]{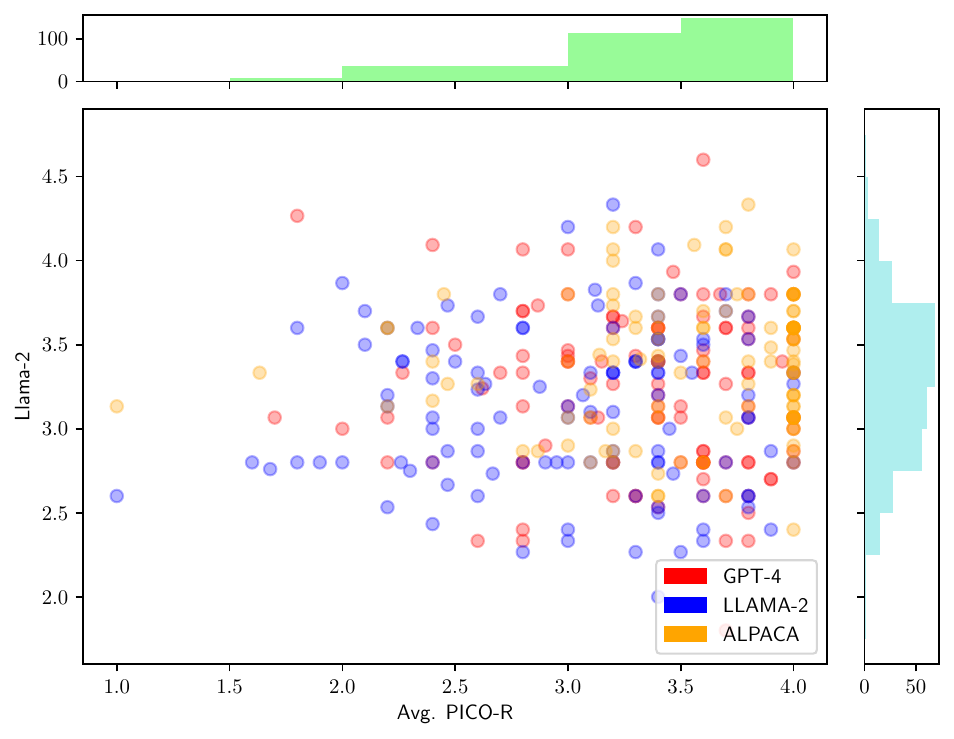}}\\
{\includegraphics[width=7.5cm, page=52]{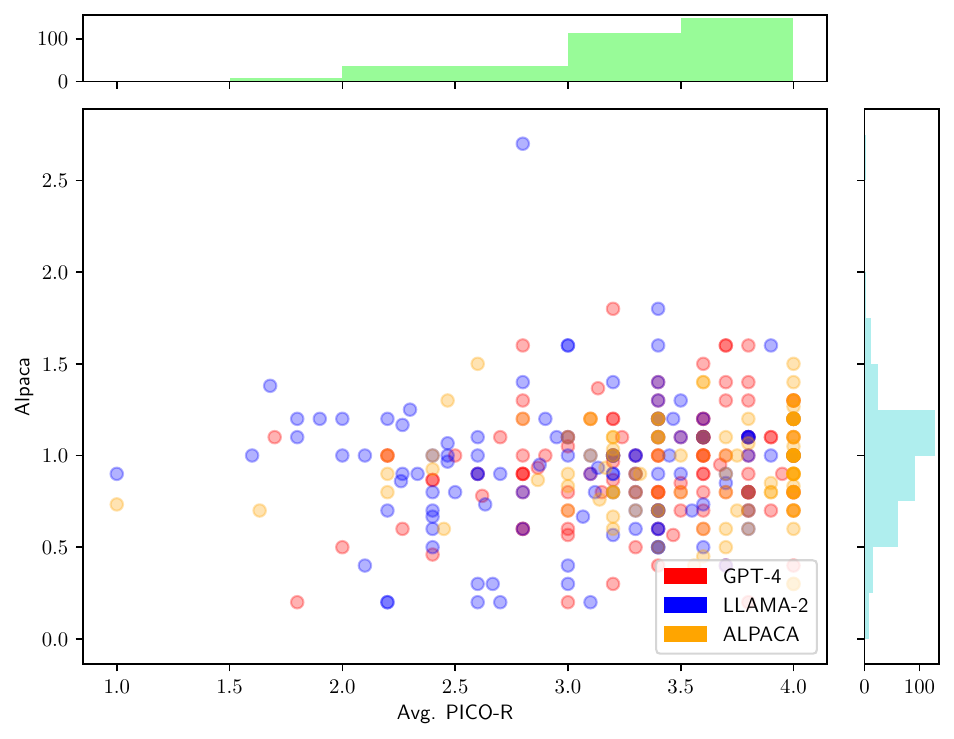}}
{\includegraphics[width=7.5cm, page=53]{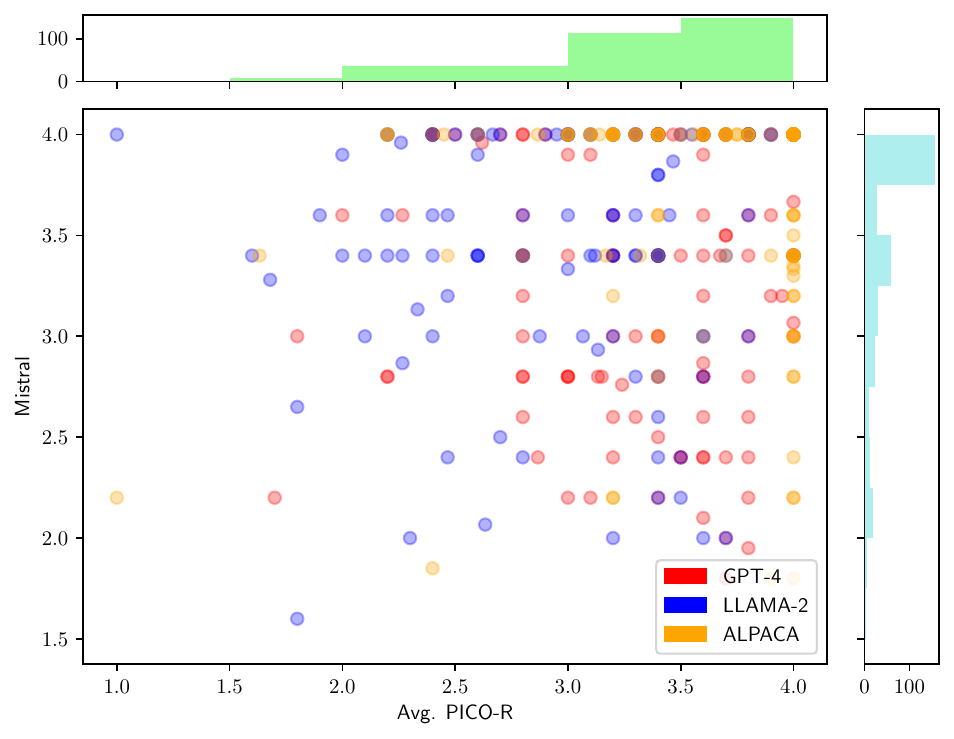}}
\caption{All traditional factuality metrics and LLMs (no Extract) plotted against avg. PICO-R. Note that human and LLMs scores are flipped (as $5-original$) to be consistent with metrics in Section~\ref{sec:factualitymodels}, hence higher is better.}\label{fig:allscatt}
\end{figure*}

Figure~\ref{fig:allscatt} shows all system evaluations plotted against human evaluations. 
Figure~\ref{fig:dist_gauss} describes the distribution of averaged {\sc FactPICO} ratings alongside those of the automatic evaluation methods. These graphs display Gaussian approximations of the standardized distributions, allowing for easy visual comparison despite differences in scales. We only observe the average of the PICO-R ratings across human and LLM evaluations.

Our primary focus in this analysis is observing how closely the distribution of automatically-derived factuality scores are to the human evaluations in {\sc FactPICO}. Upon visual inspection of Figure~\ref{fig:dist_gauss}, we see that GPT-4, Mistral, and AlignScore seem to have the most closely aligned distributions to the average human rating distribution in {\sc FactPICO}. 
In order to quantify this, we also calculated the KL divergence between the distributions of the evaluated metrics and the distribution of the averaged human ratings from {\sc FactPICO} as they are presented in Figure~\ref{fig:dist_gauss}. These results can be found in Table~\ref{table:kldiv}.
As previously hypothesized, the three distributions closest to the averaged human distribution are from AlignScore, Mistral, and GPT-4, in that order. 
Another trend that is observable in Figure~\ref{fig:dist_gauss} are the differences in the narrowness of the distributions between some LLM evaluators and dedicated factuality models. Ratings from GPT-4, Mistral, and, to a certain extent, Alpaca typically have narrow distributions, indicating an overall preference towards a single rating. On the other hand, distributions from dedicated factuality models are wider, signaling more variation.

Interestingly, these is a mismatch between well-correlated metrics and metrics whose distribution aligns closer to that of the human ratings (Figures~\ref{fig:scatt} and \ref{fig:allscatt}). Having a closely aligned distribution does not imply good correlation. Similarly, good correlation, unless it is exceptionally high, does not imply closely aligned distributions. This is the most evident in the case of AlignScore.  However, this mismatch does highlight interesting trends. A left or right skew in a score distribution references the ``strictness'' of the evaluator, with a more leaning skew indicating the evaluators are less strict and vice versa. Through this lens, the human evaluators could be viewed as not as ``strict'' compared to many of the metrics.
A possible explanation of this effect could be that the simplified nature of the text or the accurate elaborations present in the text could have been confused for factual errors.

\section{Zero-shot Prompts for LLM Evaluation}
\label{sec:pico_prompts}

Here we present the prompts used for LLM evaluation. To find the implementation details see Appendix~\ref{sec:llm_implem_details}.

\subsection{Post-processing Ratings}

The ratings produced from the prompts displayed below follow a reverse scale, where lower scores indicate the evaluated text is more factual. 
For the sake of comparison, we post-process these ratings as $5 - \textnormal{Original Rating}$, flipping the scale so that it follows the rest of the evaluated metrics.
Llama-2 is the only system that was not post-processed. We present the reasons for why this was done in Appendix~\ref{sec:llama2_analysis}.

\subsection{LLM Full-text Evaluation Prompt for PICO Elements}

The following prompt is provided as a system prompt to the large language model.

\textit{You are given an abstract and a summary. <PICOInfo> Find the <PICOElem> in accordance with PICO in both the abstract and the summary and use it rate the summary between 1 to 5.\\ \\ \\The ratings are as follows.\\ \\1 - The <PICOElem> is mentioned in the model summary and described accurately.\\2 - The <PICOElem> is mentioned in the model summary but described vaguely or somewhat inaccurately.\\3 - The <PICOElem> is mentioned in the model summary but described inaccurately or is missing critical descriptors.\\4 - The <PICOElem> is missing in the model summary.\\5 - N/A\\ \\Please provide only the rating and the rationale for the rating. Provide the rating after stating "Rating:".} \\

\noindent
The model is then queried as follows: \\

\noindent
\textit{Abstract:\\<Abstract> \\ \\Summary:\\<Summary>\\}

The tag "\textit{<PICOInfo>}" correspond to the following four descriptors depending on the evaluated PICO element. \\

\noindent
\textbf{Population:}
\textit{Population in PICO describes the type of subjects involved in the trial. Critical descriptors for population include important demographic information and any specific shared conditions.}

\noindent
\textbf{Intervention:}
\textit{Intervention in PICO describes the treatments considered in the trial.}

\noindent
\textbf{Comparator:}
\textit{Comparator in PICO describes the alternative treatment to which the intervention is being compared against.}

\noindent
\textbf{Outcome:}
\textit{Outcome in PICO describes the outcome measures used to determine results of the trial. If the primary outcome measures are not be mentioned, then the summary is critically flawed.}\\

The "\textit{<PICOElem>}" tag is replaced with the evaluated PICO element name ("population", "intervention", "comparator", "outcome").

\subsection{PICO-only LLM Evaluation Prompt}

The prompt used for PICO-only LLM evaluation is almost identical to the one above. The only change is that the first sentence of the previous prompt (\textit{"You are given an abstract and a summary."}) is changed to "\textit{You are given a list of PICO elements from an abstract and a summary.}"

\subsection{Prompt for Extracting PICO Elements}

The following prompt is used to extract PICO elements from medical text. \\

\noindent
\textit{Definition of each PICO element: \\ \\Population: The types of patients involved in the trial \\
Intervention: The treatments considered \\
Comparator: The alternative treatment to which the intervention is being compared to. \\
Outcome: What is measured. NOT what was found in the study (“result”). For example, if the study finds a drug reduces the duration of headache, the outcome here is just the “duration of headache”, not that it reduced it.  \\
\\
Identify the PICO elements in the following passage. Pull direct quotes from the passage:\\
}

\subsection{Evidence Inference Full-Text Prompt}

In {\sc FactPICO}, evaluators analyze if individual evidence inference spans from the abstract are accurately represented in the plain language summary.
The LLM evaluation is modeled after this as well, comparing evidence inference spans from the abstract to the full text of the summary.
The following is the system prompt used for this evaluation. 

\noindent
\textit{You are given a result inference span from an abstract, and you are given a summary. A result inference span corresponds to an inferred result in an experiment. Find the corresponding result inference in the summary and use it to rate the summary between 1 to 4.\\\\The ratings are as follows:\\\\1 - The result inference is mentioned and described accurately.\\2 - The result inference is mentioned but is described vaguely or is slightly inaccurate.\\3 - The result inference is critically inaccurate.\\4 - The result inference is missing in the model summary.
\\\\Please provide only the rating and the rationale for the rating. Provide the rating after stating "Rating:".\\}

\noindent
The model is then queried as follows: \\

\noindent
\textit{Result Inference Span:\\<Span from abstract> \\ \\Summary:\\<Summary>\\}

\subsection{Evidence Inference Extraction Prompts}

We used the following prompt to extract evidence inferences from the plain language summary. These extractions are subsequently compared against evidence inferences from the abstract for the evaluations. The extraction prompt is as follows: \\

\noindent
\textit{An result inference span corresponds to an inferred result in an experiment.\\\\Identify result inference spans in the following passage. Pull direct quotes from the passage:} \\

\noindent
Here is the prompt for the evaluation itself:\\

\noindent
\textit{
You are given a result inference span from both an abstract and a summary. A result span corresponds to an inferred result in an experiment. Use the result inference spans from the abstract and the summary to rate the summary between 1 to 4.\\\\The ratings are as follows:\\\\1 - The result inference is mentioned and described accurately.\\2 - The result inference is mentioned but is described vaguely or is slightly inaccurate.\\3 - The result inference is critically inaccurate.\\4 - The result inference is missing in the model summary.\\\\Please provide only the rating and the rationale for the rating. Provide the rating after stating "Rating:".
}\\

\noindent
The model is then queried as follows: \\

\noindent
\textit{Abstract:\\<Abstract Evidence Inference Span> \\ \\Summary:\\<Extracted Evidence Inferences from Summary>\\}

\section{Exhaustive Outcomes}\label{sec:exhaustive}

The outcome element in RCTs may often be represented through multiple measures, some of which may not be critical for the experiment. The omission of these non-critical outcome measures in plain language summaries usually does not impact its factuality. In {\sc FactPICO} we also asked evaluators to determine whether all outcome measures are exhaustively mentioned in the plain language summary as a separate tag {\em exhaustive}. 
This enables us to keep track of when these omissions occur without tying them to the factuality evaluation.
The addition of this annotation also enabled better agreement on outcome annotations. This was one of the factors that led to better agreement as discussed in Section~\ref{sec:discussion}. 
We also calculated agreement for exhaustive annotations through Randolph's kappa and report it to be 0.44, which signifies moderate agreement.
These collected annotations will also be included in the released data.

\section{Contradictions}\label{app:contra}
\begin{table}[h]
\centering
\small
\begin{tabular}{lrr}
\toprule
{} &  \textbf{\#C} & \textbf{\%C}\\
\midrule
ALPACA     &  \textbf{11} & \textbf{8.70} \\
GPT-4      & 16 & 10.4 \\
LLAMA-2    & 36  & 25.2 \\
\bottomrule
\end{tabular}
\caption{Total number of contradictions (\#C) and percentage of {\sc FactPICO} that is a summary with at least one contradiction (\%C).
}
\label{table:llm_analysis_contra}
\end{table}

\begin{table*}[t]
    \centering
    \scriptsize
\begin{tabular}{
p{0.97\textwidth}} \toprule
    \Example{Complete Contradiction}{Compared to OP group, the \underline{post operative diclofenac requirement was significantly less in LP group} (mean 107.14 mg) and OP group required mean of (682.35 mg).}{Llama-2}{The study found that laparoscopic pyeloplasty takes longer to do and \underline{needs more medicine to help with pain after the surgery}, but it also has fewer problems and the person stays in the hospital for less time.}{The LPA group needed LESS pain medication, not more.}\vspace{0.3em}\\
    \rowcolor{gray!10}
    \Example{Misinterpretation}{The effects of add-on prolonged-release melatonin (PRM)...were investigated in 80 patients...
In this randomized, double-blind, parallel-group study, patients were treated for 2 weeks with placebo and \underline{then randomized (1:1) to receive 2 mg of PRM or placebo nightly for 24 weeks}, followed by 2 weeks placebo.
}{Llama-2}{\underline{The medicine was given to 80 people with AD}, some of whom also had trouble sleeping.}{According to my understanding the 80 people were split 1:1 so that only 40 people were treated with melatonin and 40 received a placebo.}\vspace{0.3em}\\
    \Example{Exaggeration \textcolor{Cyan}{(fuzzy)}}{Although, laparoscopic pyeloplasty has the disadvantages of longer operative time and requires significant skill of intracorporeal knotting \underline{but it is here to stay and represents an emerging standard of care.}}{GPT-4}{So, they think the \underline{camera surgery is becoming the best choice even if it's a bit more difficult to do.}}{This is not necessarily linked with each other.} \\
    \bottomrule
\end{tabular}
\vspace{-0.5em}
\caption{Examples of contradictions found in plain language summaries within {\sc FactPICO}.}
\label{fig:span_ex_contra}
\vspace{-0.5em}
\end{table*}
We define \emph{Contradictions} as pieces of information in the plain language summary that meaningfully disagree with the input abstract.
In addition to the characteristics described in the main paper, we also ask evaluators to find spans in both the original abstract and the plain language summary that contradict each other and provide a rationale. 
The results are presented in Table~\ref{table:llm_analysis_contra}, following similar trends to that found for other dimensions in Table~\ref{table:llm_analysis}.

We observed poor agreement overall for this question, largely due to ``fuzzy'' nature of this task.
We define fuzziness as the difficulty to exhaustive distinguish such characteristics within a text.
Table~\ref{fig:span_ex_contra} shows a fuzzy, contradiction example which demonstrates how certain exaggerations may not be detected by the reader because it may be considered acceptable to use as part of the simplified language style. Annotators with different levels of discernment will likely suffer from low agreement for these types of annotations due to this phenomenon \cite{Trienes:2024:arXiv}.
These collected annotations will also be included in the released data.

\section{Llama-2 Correlation}
The negative correlations under the Flip column of Table \ref{table:llama_flip} demonstrate the worser performance of Llama-2 scores that are flipped (post-processed to be 5-x) compared to its original ratings. Out of all LLM evaluators, this occurrence is unique to Llama-2 . We attribute this to Llama-2's struggles to remain consistent and faithful to the rating system defined in the prompt. Through analyzing its rationales, we discovered that Llama-2 often ranks summaries it explains as accurate with high numbers, which was opposite to the instructions.

This Llama-2 rationale assigns a rating of 4 to a summary that it describes as accurate and sufficiently detailed, which is the opposite of the rating system provided in the prompt.
\vspace{-0.5em}
\begin{myquote}{0.1in}
\small
 "The population in the abstract and summary is described accurately and with sufficient detail. The authors mention the type of subjects involved in the trial, including their demographics (age, sex, and duration of HIV infection) and any specific shared conditions (abacavir use). The summary provides a brief overview of the study population, including their baseline characteristics and the results of the intervention.The summary accurately reflects the information provided in the abstract and provides a clear and concise summary of the study population. The authors use descriptive language to highlight the key findings related to the study population, such as the slight improvement in endothelial function after maraviroc intensification. The summary does not contain any inaccuracies or vagueness that could affect the reader's understanding of the study population. \textbf{Therefore, I would rate the summary as a 4, indicating that the population is described accurately and with sufficient detail.}"
\vspace{-0.5em}
\end{myquote}

Additionally, the following rationale demonstrates the hallucinations that Llama-2 makes about the rating scale, as it assumes a ranking order of 1-5 with 5 being the most accurate. Not only is 5 out of the range of the rating options given in the prompt, but also this scale is inconsistent with Llama-2's other evaluation scale of using 4 as its highest rating.

\begin{myquote}{0.1in}
\small
"The population in the abstract and summary is described as "drug users" or "people using illicit opiates," which is a broad and vague term that does not provide critical demographic information. The summary does not provide enough detail about the specific characteristics of the population, such as age, gender, ethnicity, or comorbidities. Additionally, the summary does not indicate whether the population was selected randomly or if there were any exclusion criteria.\\ \\To rate the summary, I would give it a score of 4 because it does not provide sufficient information about the population involved in the study. \textbf{A rating of 5 would indicate that the summary provides detailed and accurate information about the population, including critical demographic information and any specific shared conditions.}"
\vspace{-0.5em}
\end{myquote}

This analysis explains why non-flipped Llama-2 ratings were more correlated with human evaluation. Llama-2 has a tendency to not follow rating instructions and rate more accurate summaries with higher numbers, defeating the purpose of flipping its ratings in post-processing. Furthermore, Table \ref{table:grade} shows how Llama-2 (non-flipped) is capable of rating plain language summaries similarly to human annotators in Table~\ref{table:llm_analysis}. If we flipped Llama-2 ratings, the average ratings for Llama-2 would have given opposite results. Therefore, we report the non-flipped Llama-2 evaluations throughout this paper to most accurately depict Llama-2's intended ratings.
\label{sec:llama2_analysis}
\begin{table}
\small
\centering
\begin{tabular}{lrr}
\toprule
{} &  \textbf{Flip} &  \textbf{No Flip} \\
\midrule
Population   &       -0.047 &        0.042 \\
Intervention &       -0.063 &        0.059 \\
Comparator   &       -0.038 &        0.036 \\
Outcome      &       -0.082 &        0.078 \\
Results      &       -0.058 &        0.056 \\
Avg. PICO-R  &       -0.060 &        0.055 \\
\bottomrule
\end{tabular}
\caption{Kendall's Tau correlations between {\sc FactPICO} ratings and the flipped and non-flipped LLAMA-2 ratings.}
\label{table:llama_flip}
\end{table}

\section{Rationale Length Analysis}
\begin{table}[t]
\centering
\small
\begin{tabular}{lrrr}
\toprule
{} &  \textbf{GPT-4} & \textbf{Llama-2} & \textbf{Alpaca}\\
\midrule
\textbf{P}     &  66.6 &  136.9 &  33.3 \\
\textbf{I}     &  64.7 &  121.4 &  30.0 \\
\textbf{C}    &  82.5 &  140.2 &  30.1 \\
\textbf{O}   &  73.8 &  133.6 &  28.2 \\
\textbf{R}   &  63.0 &   90.8 &  45.2 \\
\midrule
{} & \textbf{Mistral} & \textbf{Extract} & \textbf{Human} \\
\midrule
\textbf{P}     &  66.2 &  65.1 & 15.4\\
\textbf{I}      &  56.9 &  55.4 & 15.0\\
\textbf{C}    &  83.4 &  81.0 & 12.7\\
\textbf{O}   &  66.5 &  65.4 & 24.1\\
\textbf{R}   &  71.0 &  62.5  & 13.6\\
\bottomrule
\end{tabular}
\caption{Average number of tokens for rationales from all systems. Human represents rationales in {\sc FactPICO}.}
\label{table:rat_lengths}
\end{table}

\label{sec:rat_length}

Table~\ref{table:rat_lengths} shows the average number of tokens for rationales in {\sc FactPICO} and those generated by LLMs. As discussed in Section~\ref{sec:prelim_rat}, Llama-2 produced the longest rationales. GPT-4, Mistral, and GPT-4 pipelined with PICO-R extraction generated rationales with similar lengths. Among the LLMs, Alpaca produced the shortest rationales. However, overall, expert-written rationales in {\sc FactPICO} have the shortest lengths. This is largely because evaluators tend to justify themselves as concisely as possible, especially for easy evaluation instances, such as when an element is clearly mentioned accurately or clearly missing.

\section{LLM Rationale Errors}
\label{sec:rat_errors}

Table~\ref{fig:llm_rat} shows several examples of erronous rationales generated by various LLMs. 
These rationales illustrate several patterns of errors exhibited when these systems generate rationales.
The examples from Mistral show the tendency to ``forget'' to evaluate the summary as described in Section~\ref{sec:prelim_rat}. Mistral here either completely ignores the input summary or it passes off text from the abstract as belonging to the summary, as exemplified by the very first example. In addition to this issue, we also observe that Mistral generates illogical and contradictory language in its rationales (ex 3).

The examples from Llama-2 exemplify the length of its rationales as well as its tendency to use irrelevant ideas in its justifications. For example, it may acknowledge factual information from the summary (\emph{``While the summary mentions the age range of the patients
(adults) and the type of surgery (elective)''}), but draws invalid conclusions (\emph{``it does not provide additional demographic information''}).

Rationale examples from GPT-4 here ignores critical factual errors in the summary. For example, the first GPT-4 example shows that the model acknowledges that the summary left critical characteristics missing (\emph{``Although the summary doesn’t specifically mention `adult' patients or the type of hospital''}), but claims that \emph{``this doesn't seem to be crucial information''}, downplaying its effect during evaluation. 

We also included some examples from Alpaca, showing its rationales tend to be very similar to the the rating descriptions provided through the prompt, and exceptions to this tend be nonsensical and illogical text.

\newcommand{\Examplee}[3]{
    \textbf{Human Rationale:} #1 \newline
    \textbf{LLM Rationale [#2]:} #3
}
\begin{table*}[t]
    \centering
    \scriptsize
\begin{tabular}{
p{0.97\textwidth}} \toprule
    \Examplee{The population of the study is not mentioned clearly at all.}{Mistral}{
       The population in the abstract and the summary is described accurately. The population includes 16 eyes of 16 patients with corneal neovascularization (NV) secondary to herpetic keratitis, graft rejection, chemical burn, pemphigoid, and recurrent ulcer. The patients were randomly assigned to receive 2.5 mg of bevacizumab or 1 mg of ranibizumab through subconjunctival and intrastromal routes. The researchers followed up with the patients for one month after the injections and calculated the corneal NV areas using Image J software. The results showed that bevacizumab injection resulted in a more effective and stable regression of corneal NV compared to ranibizumab injection.
    } \\
    \vspace{0.05cm}\\
    \rowcolor{gray!10}
    \Examplee{It is mentioned that the patients included all (here called vaguely "people") had a kidney transplant, the size of population is missing.}{Mistral}{The population in PICO is mentioned in the abstract and is described accurately. The population includes 90 patients who were prepared for renal transplantation during one year and were randomly classified into two groups: one group with a stent attached to a Foley catheter and the other group with a conventional technique (stent separated from Foley). The population is described in terms of the study design, which is a randomized controlled trial.} \\
    \vspace{0.05cm}\\
    \Examplee{Outcome (Low‐density lipoprotein cholesterol (LDL‐C), PCSK9, and alirocumab levels) is mentioned partly. Missing: PCSK9}{Mistral}{The outcome measures used in this study are not mentioned in the summary. The summary describes the results of the study, including the reduction in LDL-C levels and the duration of effect, but it does not provide specific information on the outcome measures used to determine these results.}\\
    \midrule
    \rowcolor{gray!10}
    \Examplee{Comparator (placebo) is not mentioned.}{Llama-2}{The comparator in the PICO model is the "placebo" condition, which is the control group against which the intervention (THC, CBD, or their combination) is being compared. In the abstract, the comparator is mentioned accurately, but the description is somewhat vague. For example, the authors state that CBD "improves recognition of emotional facial affect," but they do not provide specific details on how this improvement occurs or how it compares to the placebo condition. In the summary, the comparator is mentioned accurately, but the description is still somewhat vague. The authors state that THC alone impairs recognition, while CBD improves it, but they do not provide specific details on the magnitude of these effects or how they compare to the placebo condition. Additionally, the authors mention that the combination of THC and CBD had no effect, but they do not provide any specific data to support this claim.Overall, while the comparator is mentioned accurately in both the abstract and summary, the descriptions provided are not detailed enough to fully understand the results of the study or to compare the interventions being tested. }\\
    \vspace{0.01cm}\\
    \Examplee{Population (200 pat., adults) is mentioned and described accurately.}{Llama-2}{The population in the summary is described somewhat inaccurately. While the summary mentions the age range of the patients (adults) and the type of surgery (elective), it does not provide additional demographic information such as gender, race, or socioeconomic status. Additionally, the summary does not mention any specific shared conditions that might be relevant to the study.The summary does provide some critical descriptors, such as the types of anesthesia used (separate written consent for anesthesia) and the outcome measures used to assess the efficacy of the informed consent process (adequacy of consent index). However, the summary could benefit from more detailed information about the patient population, such as the number of patients in each group and the duration of follow-up. Overall, while the summary provides some useful information about the study population, it could be improved by providing more detailed demographic information and critical descriptors.}\\
    \vspace{0.01cm}\\
    \rowcolor{gray!10}
    \Examplee{The population is described almost correctly as "60 healthy adults who are a little overweight". Their diagnose is not "a little" overweight but overweight.}{Llama-2}{The summary provides some information about the population in accordance with PICO, but it is missing critical descriptors in some cases. For example, the summary does not mention the age range of the participants, which is an important demographic characteristic in this study. Additionally, the summary does not provide details about the specific shared conditions among the participants, such as the level of obesity or the presence of metabolic syndrome. While the summary mentions the type of subjects involved in the trial (healthy adults), it does not provide enough detail about the population to accurately rate the summary. Therefore, I will give a rating of 4.} \\
    \midrule
    \Examplee{Population (pat.) is mentioned- yet information about demographics are missing (adults, number of pat.).}{GPT-4}{
    The population is accurately described in the summary as patients who are going to undergo surgery. This matches the abstract which specifies the population as adult patients undergoing elective surgery at a tertiary care teaching hospital. Although the summary doesn't specifically mention "adult" patients or the type of hospital, this doesn't seem to be crucial information in this context as the aim of the study is more focused on the procedure (anesthesia and consent process) rather than population characteristics. So, the description in the summary is considered accurate.
    } \\
    \vspace{0.01cm}\\
    \rowcolor{gray!10}
    \Examplee{The intervention is mentioned vaguely cause it says "one group got the medicine before surgery " - this could refer to any medicine... only in the introduction ketamine is mentioned, yet also it is not made clear that it was an injection of a small doses of Ketamine that was given as intervention.}{GPT-4}{The abstract and the summary both document the intervention accurately. The intervention mentioned in the abstract is the preemptive use of ketamine before surgery to reduce postoperative pain. The summary accurately describes the same by mentioning that one group got the medicine (ketamine) before surgery, which helped to lessen the pain after the surgery. Both the dosages and the routes of administration have been accurately described in the summary, in line with the abstract's details. The outcomes have also been similarly explained in both. Hence, the summary is rated as 1, indicating that the interventions are mentioned and described accurately.} \\
    \vspace{0.01cm}\\
    \Examplee{Comparator (sham tDCS) is not mentioned and missing.}{GPT-4 Extract}{The summary mentions a form of comparator, represented as "when the other area of the brain called the DLFPC was stimulated in a different way", but it's vaguely related to the original abstract's Comparator of "sham tDCS". The summary fails to accurately explain that the comparison was against a fake (sham) treatment. Therefore, it can be seen as being inaccurately depicted in the summary.} \\
    \midrule
    \rowcolor{gray!10}    
    \Examplee{Intervention (MIC, NPC) is mentioned and described accurately.}{Alpaca}{The interventions are mentioned in the model summary but described inaccurately.} \\
    \vspace{0.01cm}\\
    \Examplee{The comparator is correctly mentioned as "inpatient vaginal PGE2" IOL.}{Alpaca}{The comparator is described accurately.}\\
    \vspace{0.01cm}\\
    \rowcolor{gray!10}
    \Examplee{The intervention is correctly described as techniques for removing ureteral stents, linked stent technique.}{Alpaca}{The linked stent technique has a lower complication rate than the separate stent technique, suggesting its effectiveness and success.} \\
    \bottomrule
\end{tabular}
\caption{Examples of erroneous rationales generated by LLMs compared against rationales in {\sc FactPICO}.}
\label{fig:llm_rat}
\end{table*}

\section{Full Text of Abstract in Figure~\ref{fig:main_fig2}}
\label{sec:abs_full_txt}

Due to space constraints, Figure~\ref{fig:main_fig2} substituted the full abstract used during evaluation with a link. The full text of this abstract is included below:

\begin{myquote}{0.1in}
\small
NicVAX®, a nicotine vaccine (3’AmNic-rEPA), has been clinically evaluated to determine if higher antibody concentrations are associated with higher smoking abstinence rates and if doses and frequency of administration are associated with increased antibody response. This randomized, double-blinded, placebo-controlled multicenter clinical trial (N=301 smokers) tested 200 and 400 µg doses administered 4 or 5 times over 6 months compared to placebo. 3’AmNic-rEPA recipients with the highest serum anti-nicotine antibody response (top 30\% by AUC) were significantly more likely to attain 8 weeks continuous abstinence from weeks 19 through 26 than the placebo recipients (24.6\% vs. 12.0\%, p=0.024, OR=2.69, 95\% CI, 1.14–6.37). The 5 injection 400 µg dose regimen had the greatest antibody response and had significantly higher abstinence rates than placebo. This study demonstrates proof-of-concept that 3’AmNic-rEPA elicits antibodies to nicotine and is associated with higher continuous abstinence rates, justifying its further development as a treatment for nicotine dependence.
    
\end{myquote}

\onecolumn

\section{Notable Examples}\label{app:factpicoexamples}

In this section, we show several examples of interest from {\sc FactPICO}. We present relevant snippets from the abstract and the appendix and show the rating and rationale given by evaluators.

\subsection{Obvious Errors}

In the following table, we show examples of factual errors that were easily identified by the expert evaluators. They stood out as clear examples of the summary misrepresenting the information in the abstract.

\newcommand{\examp}[5]{
    #1 & #2 & #3 & #4 & #5 \\
}

\begin{small}
\begin{longtable}{l p{3.1cm} p{3.1cm} p{3.1cm} p{3.1cm}}
    \toprule
    \textbf{System} & \textbf{Abstract} & \textbf{Summary} & \textbf{Question Type + Rating} & \textbf{Rationale}\\ \midrule \endhead
    \examp{Alpaca}{
    Patients with inoperable or persistent/recurrent CTEPH (n=261; mean± SD age 59±14 years; 66\% women) were randomised to riociguat (up to 2.5 mg three times daily) or placebo.
    }{
        We compared the effects of riociguat on haemodynamics in patients with inoperable chronic thromboembolic pulmonary hypertension (CTEPH) or persistent/recurrent CTEPH after pulmonary endarterectomy.
    }{
        Comparator \newline Missing 
    }{
        comparator (placebo) is not mentioned
    }\midrule
    \examp{Alpaca}{
        reatment-naive, non-cirrhotic patients with chronic hepatitis C genotype 6 were treated with pegylated interferon $\alpha$-2a (180 $\mu$g/week) and ribavirin (800–1,200 mg, according to weight) for 4 weeks
    }{
        The study included 152 patients who had an RVR (rapid virological response) and were randomly assigned to either a 24-week or 48-week treatment regimen.
    }{
        Intervention \newline Missing critical \newline information 
    }{
        The intervention misses the applied medication which is a critical descriptor as the mentioned "treatment regimes" could be anything.
    }\midrule
    \examp{Llama-2}{
        In Group A, 1 $\mu$g/kg of dexmedetomidine was intravenously administered for 10 min, followed by the maintenance infusion of dexmedetomidine 0.2 $\mu$g/kg/hr after 5 min of intrathecal bupivacaine injection. Patients in Group B received same dose of dexmedetomidine by intravenous administration before 5 min of intrathecal bupivacaine injection.
    }{
        This study looked at how two different ways of giving a medicine called dexmedetomidine before a spinal anesthesia.
    }{
        Intervention \newline 
        Described \newline inaccurately 
    }{
        Intervention (timing: dexamedetomidine after spinal anesthesia) is mentioned inaccurately. The original refers to an application before and after spinal anesthesia. But the summary mentions that TWO different ways of giving dexmedetomidine BEFORE spinal anesthesia were tested.
    }\midrule
    \examp{GPT-4}{
        A total of 84 patients with ureteral calculi 5 to 10 mm in diameter were divided into two groups.
    }{
        They studied two groups of people who all had this treatment.
    }{
        Population \newline
        Missing critical \newline information 
    }{
        The summary only mentiones "They studied two groups of people who all had this treatment". The diagnose is missing, the fact that these are patients and not just people and the size of the group is missing.
    }
    
\bottomrule
\caption{}\label{table:fact_obvious}\\
\end{longtable}
\end{small}

\subsection{Subtle Errors}

In the following table, we show examples where the error is harder to detect at a first glance. These errors disguise themselves as appearing factual, but a closer reading of the text would reveal factual inconsistencies.

\begin{small}
\begin{longtable}{l p{3.1cm} p{3.1cm} p{3.1cm} p{3.1cm}}
    \toprule
    \textbf{System} & \textbf{Abstract} & \textbf{Summary} & \textbf{Question Type + Rating} & \textbf{Rationale}\\ \midrule \endhead
    \examp{GPT-4}{The sample comprised of forty patients of both sexes diagnosed as acute and transient psychotic disorder who were randomly assigned to high dose (20 mg/day) and low dose (5 mg /day) haloperidol groups with equal number of subjects (n=20) in both groups.}{This passage is about a study where they wanted to see if giving bigger or smaller amounts of a medicine(called haloperidol) would work better to help people who had a certain type of brain sickness
}{
    Intervention \newline Described vaguely 
}{
    The intervention is vaguely described as smaller doses from haloperidol, which could be any doses - not even making clear that it is in itself a "low doses".
}\midrule
    \examp{GPT-4}{
        Seventy-four patients with American Society of Anesthesiologists physical status classification I or II were randomly allocated into two groups.
    }{
        In this experiment, they divided patients into two groups. 
    }{
        Population \newline Missing critical \newline descriptors 
    }{
        The population (patients) is mentioned. Nevertheless, the specific condition of having an ASA score I / II is not mentioned, which means that the patients are rather healthy and the doctors do not expect so many complications..
    }\midrule
    \examp{GPT-4}{
        Two variants near IRS1, rs1522813 and rs2943641, were genotyped in 738 overweight/obese adults (age 60 ± 9 years; BMI 32.7 ± 3.9 kg/m2) randomly assigned to one of four weight-loss diets (a deficit of 750 kcal/day of caloric intake from baseline) varying in macronutrient contents for 2 years
    }{
        They looked at individuals who were overweight and split them into two groups
    }{
        Population \newline Described vaguely 
    }{
        The summary only mentions the population as "individuals who were overweight", different facts from the abstract are missing like their genotype, age and BMI.
    }\midrule
    \examp{GPT-4}{
        Group 1 consisted of 14 listeners (number of females=11) who were 41–61 years old [mean=53.18; standard deviation (SD)=5.97].
    }{
        They tested two groups of adults, one younger and one older, using a special brain test.
    }{
        Comparator \newline
        Described vaguely 
    }{
        Comparator (participants aged 41- 61 and their performance on the WCJ-III test) is mentioned vaguely as younger people doing a special brain test, which leaves too much space for interpretations.
    }\midrule
    \examp{Llama-2}{
        At day 56, there were more recurrent infections in the chloroquine arm (8.9\%, 95\% CI 6.0-13.1\%) than the dihydroartemisinin-piperaquine arm (2.8\%, 95\% CI 1.4-5.8\%), a difference in cumulative recurrence rate of 6.1\% (2-sided 90\% CI +2.6 to +9.7\%). The log-rank test comparing the survival curves confirmed the superiority of dihydroartemisinin-piperaquine over chloroquine (p = 0.003).
    }{
        The study found that dihydroartemisinin-piperaquine is better than chloroquine at preventing malaria from coming back after treatment
    }{
        Outcome \newline Missing critical 
        \newline information 
    }{
        Only recurrence of infection is mentioned as Outcome, other more important like side effects or time till recovery and overall survival are not mentioned.
    }
\bottomrule
\caption{}\label{table:fact_subtle_error}\\
\end{longtable}
\end{small}

\subsection{Edge Cases}

Here we show examples that evaluators felt were "edge cases." In other words, there is some subjectivity involved in rating these examples according to the {\sc FactPICO} framework.

\begin{small}
\begin{longtable}{l p{3.1cm} p{3.1cm} p{3.1cm} p{3.1cm}}
    \toprule
    \textbf{System} & \textbf{Abstract} & \textbf{Summary} & \textbf{Question Type + Rating} & \textbf{Rationale}\\ \midrule \endhead
    \examp{Llama-2}{
        We compared the haemodynamic effects of riociguat in patients with inoperable chronic thromboembolic pulmonary hypertension (CTEPH) or persistent/recurrent CTEPH 
    }{
        This is a study about a medicine called riociguat. \ul{It helps people with a bad heart.}
    }{
        Addition \newline Factual
    }{
        in first instance CTEPH is a lung disease, that can affect the heart and can cause a "bad heart" as a result.
    }\midrule
    \examp{GPT-4}{
        adult patients, scheduled for elective cardiac surgery, were divided into three groups: saline, lignocaine (20 mg) and methylprednisolone 
    }{
        They tried mixing the medicine with either saltwater
    }{
        Comparator \newline Described vaguely 
    }{
        the comparator saline is described vaguely as saltwater.
    }\midrule
    \examp{GPT-4}{
        Changes in the quality of life measured with EQ-5D in the group of men who carried out the PAPP were statistically significant when comparing between groups. However changes in cardiopulmonary function were not as relevant when comparing between groups.
    }{
        After three months, they found that the men who did the exercise program felt better although there was no change in how they breathed or how their hearts worked. The women did not feel any better. 
    }{
        Outcome \newline Described vaguely 
    }{
        To me the outcome "felt better" as described in the summary seems vague cause it leaves open what a good feeling actually is in this context. The abstract states clearly that it is "quality of life evaluated by a valid questionnaire" - I choose vaguely for this reason.
    }\midrule
    \examp{GPT-4}{
        Intra-articular morphine was as effective as intra-articular triamcinolone for analgesia in patients with osteoarthritis knee. Intra-articular morphine is possibly a better option than intra-articular steroid as it has lesser side effects.
    }
    {
        \ul{So, if granny or grandpa is having pain because of osteoarthritis, an injection with morphine could help them better.}   
    }{
        Addition \newline Non-factual
    }{
        This is definitely not factual or good advice to give, even though the information within is technically true.   
    }
\bottomrule
\caption{}\label{table:fact_edge_cases}\\
\end{longtable}
\end{small}

\section{Evaluation Interface}\label{app:interface}
{\sc FactPICO} evaluations were collected through the Thresh platform \citep{heineman2023thresh}. This platform allows for the creation of a customizable interface for any text generation task, supporting a variety of question types and text span highlighting. The platform is hosted as a website and the customizable interface is encoded in YAML. 

We modify the original Thresh interface to support integration with the Google Drive API.
\footnote{This version of Thresh is available as a fork of the original project through this link: \url{https://github.com/SebaJoe/thresh}} 
This integration enables annotators to securely retrieve their saved evaluations while having a copy of their evaluations stored in a shared Google drive. This facilitates a straightforward organization of evaluation files in an easily accessible, shared location. 
\label{sec:interface}

\begin{figure}
    \centering
    \includegraphics[width=\linewidth]{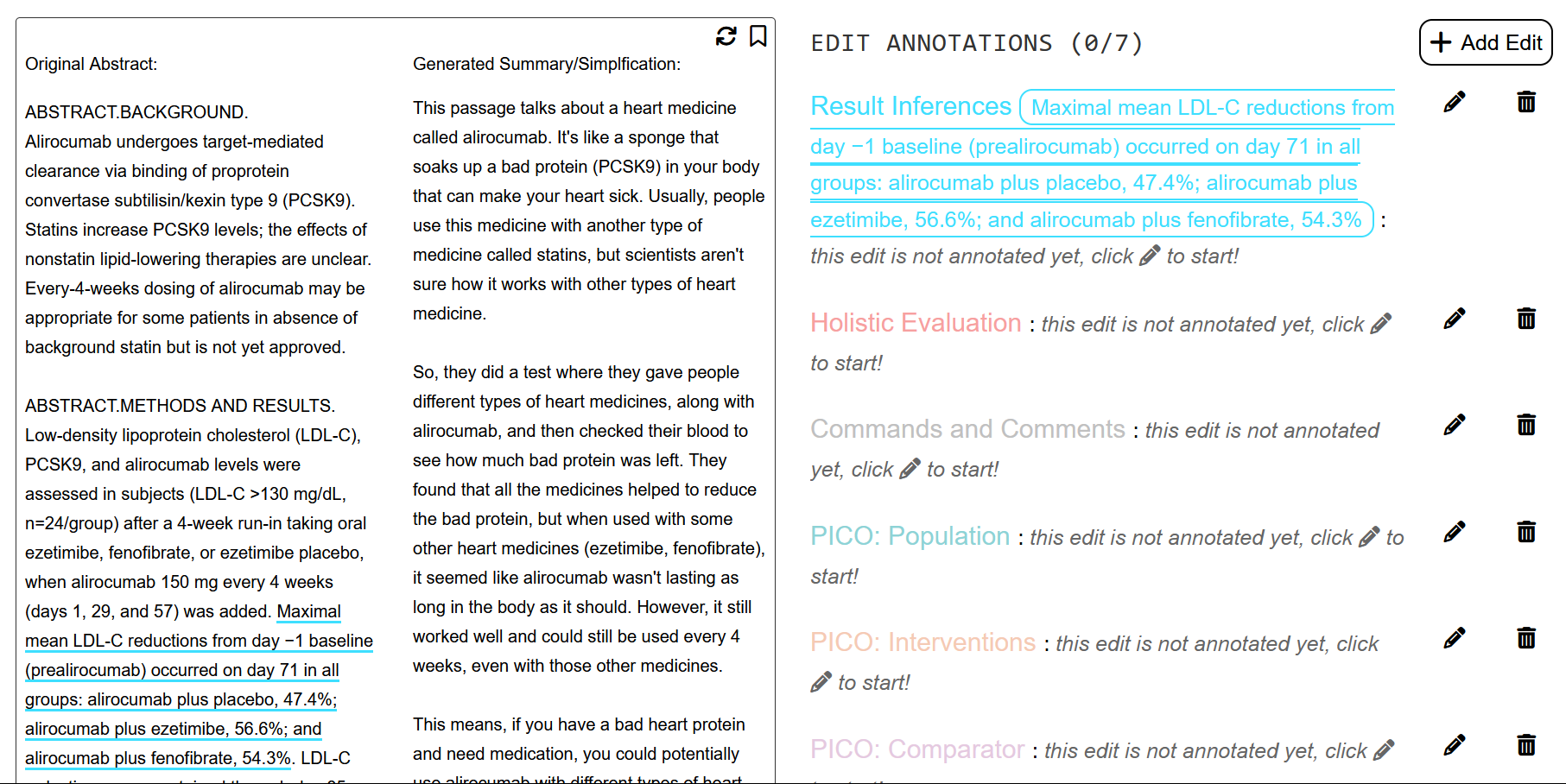}
    \includegraphics[width=\linewidth]{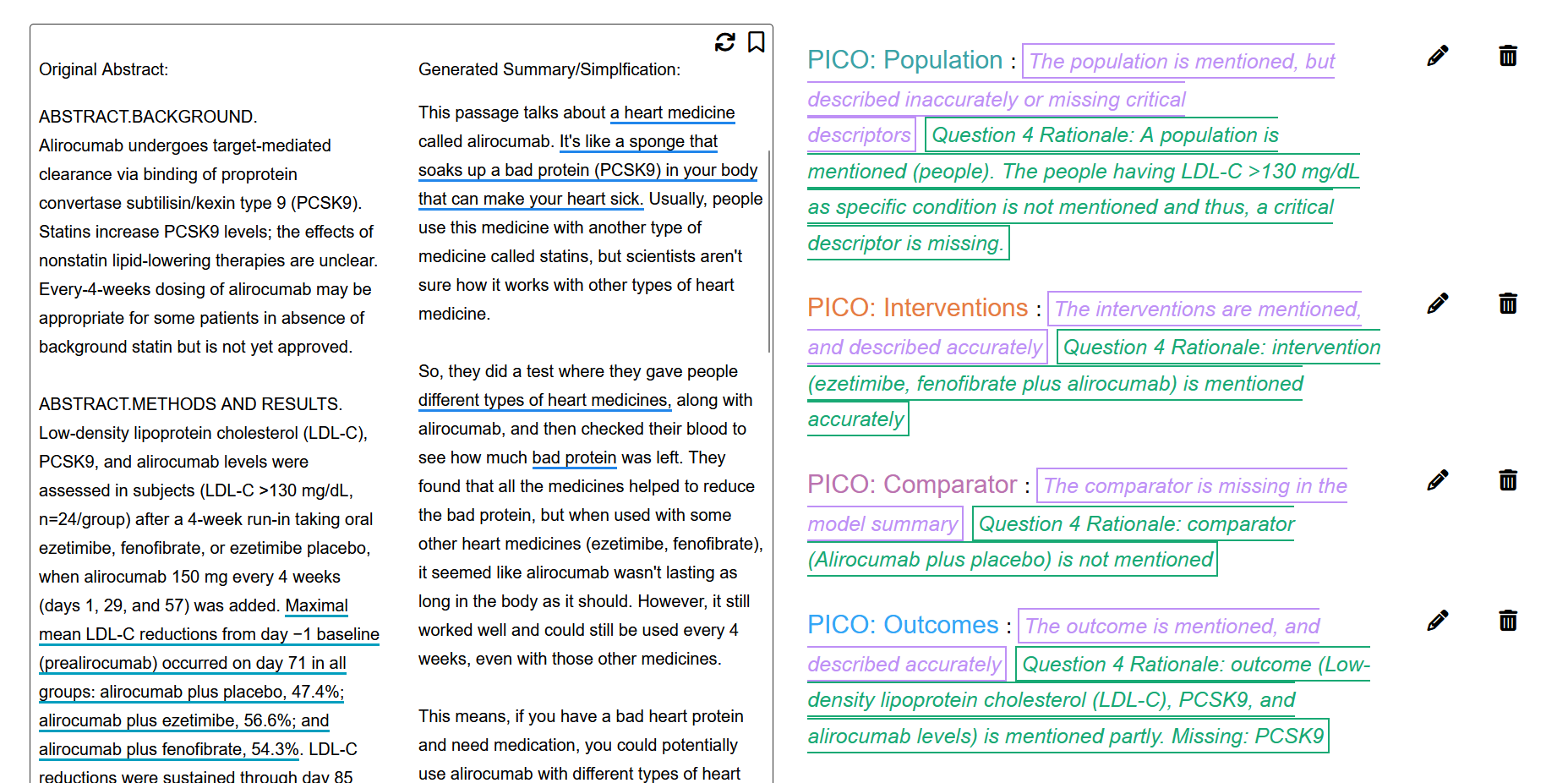}
    \caption{The initial state of the Thresh interface (top) and the state after annotations have been completed (bottom).}
    \label{fig:inter12}
\end{figure}

\begin{figure}
    \centering
    \includegraphics[width=\linewidth]{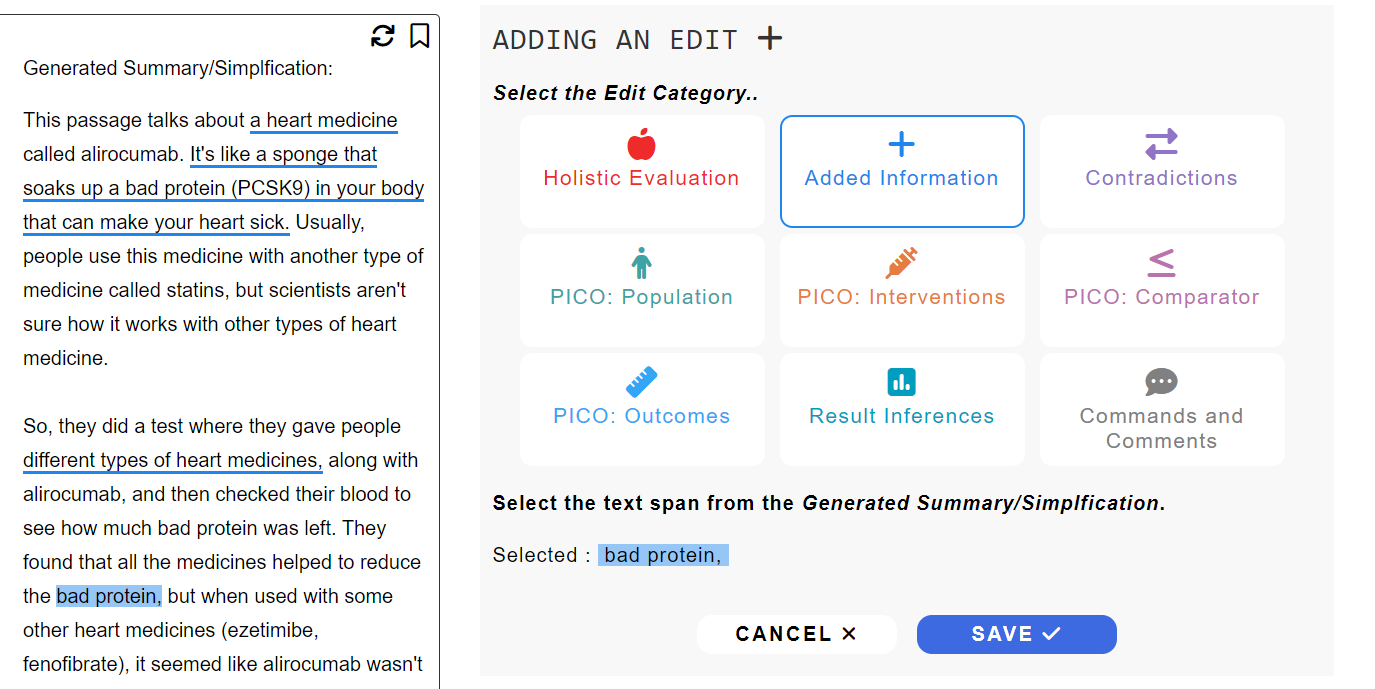}
    \caption{Annotating an added information span in a plain language summary.}
    \label{fig:inter3}
\end{figure}

\begin{figure}
    \centering
    \includegraphics[width=0.6\linewidth]{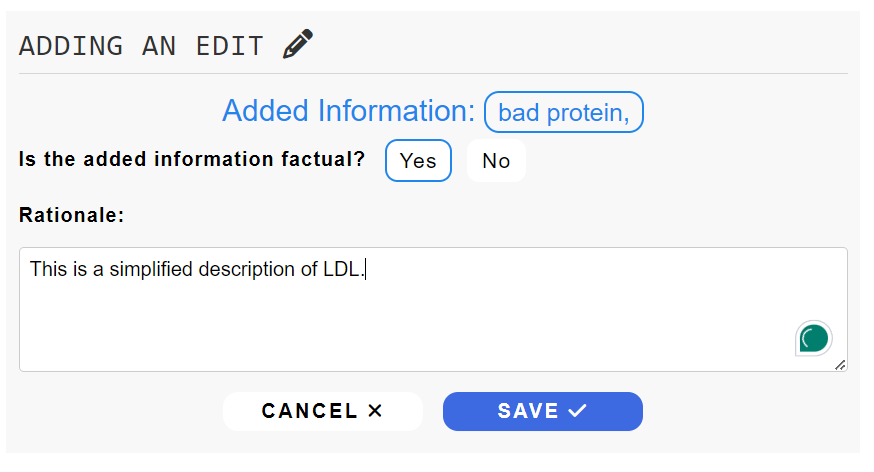}
    \includegraphics[width=0.6\linewidth]{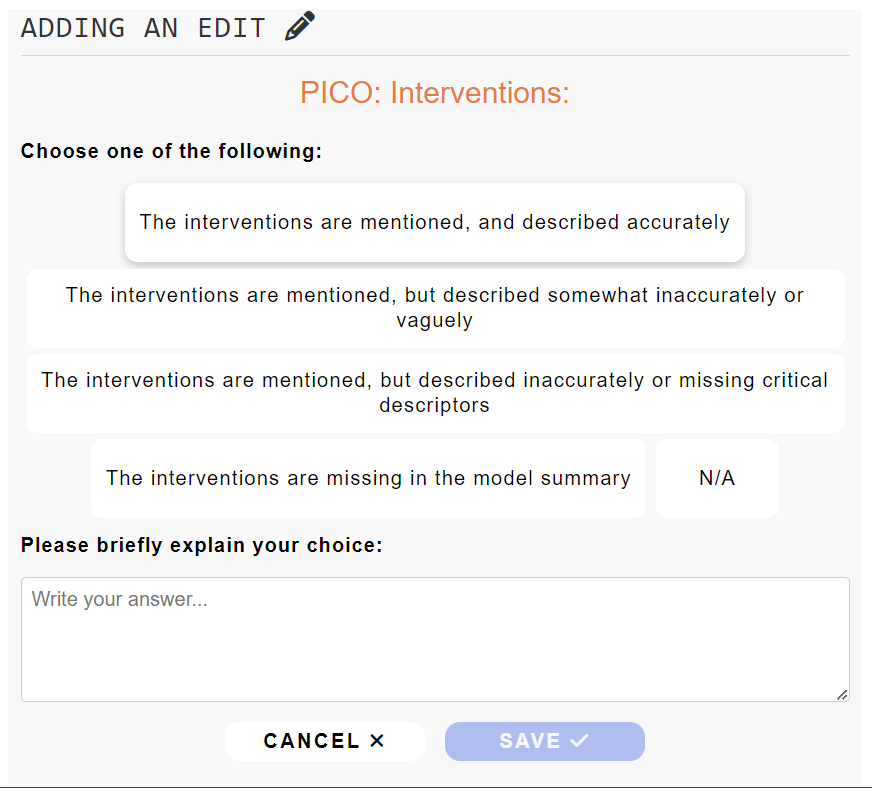}
    \caption{Interface for answering questions regarding added information (top) and PICO interventions (bottom).}
    \label{fig:inter45}
\end{figure}

Figure~\ref{fig:inter12} shows both the initial state of the interface and the state after annotations have been completed.
The initial state shows almost all questions evaluators would have to answer as "edit annotations." However, span-level annotations have to be created by adding an edit, selecting the type of edit (i.e. Added Information), and highlighting text.
Figure~\ref{fig:inter3} shows how evaluators can select spans in text to evaluate.
Evaluators answer questions by clicking the pencil icon next to each edit.
Figure~\ref{fig:inter45} shows how questions are displayed for evaluators to answer.

\end{document}